\documentclass[10pt]{article} 
\usepackage[preprint]{tmlr}

\usepackage{hyperref}
\usepackage{url}
\usepackage{booktabs}
\usepackage{graphicx}
\usepackage{multirow}

\title{Scoring-Rule Sensitivity in LLM Evaluation:\\
Evidence from Indian Financial Regulatory Text}

\author{\name Rajveer Singh Pall \email rajveerpall04@gmail.com \\
\addr Department of Computer Science and Business Systems\\
Gyan Ganga Institute of Technology and Sciences, Jabalpur, India}

\begin{document}

\maketitle

\begin{abstract}
A benchmark leaderboard reports one accuracy number, but that number is a joint product of a model
and a scoring rule. We ask what changes when only the rule changes, and find that the ordering of
evaluated systems is highly sensitive to it. IndiaFinBench, introduced here, is a 406-item
expert-annotated benchmark over SEBI and RBI regulatory text covering regulatory interpretation,
numerical reasoning, contradiction detection, and temporal reasoning; it is built on
jurisdiction-specific regulatory text uncommon in Western financial benchmarks, which lowers but
does not rule out the chance that performance reflects prior familiarity with this specific material.
We score twelve contemporary LLMs, evaluated zero-shot, under two primary scoring regimes over the
same predictions---strict four-stage string matching, and the verdicts of a cross-model judge
sharing no model family with any evaluated system---plus one operational composite regime
(strict-or-judge, included to quantify asymmetric judge application rather than proposed as
authoritative).

The two primary regimes show no positive rank correspondence: Spearman $\rho = -0.273$
between strict and judge-only accuracy across the twelve models ($p = 0.39$, not significant), and no model holds
its rank across all three regimes. Strict accuracy spans 75.12--89.66\% (Gemini~2.5~Flash highest,
DeepSeek-R1-Distill-Llama-70B lowest); judge-only accuracy spans 87.19--94.83\% with a different
leader and trailer (Gemini~2.5~Pro highest, Gemma~3~4B lowest); the composite regime compresses
the field further, to 94.1--98.0\%, under which DeepSeek-R1-Distill-Llama-70B and LLaMA-3.3-70B tie
for first despite DeepSeek ranking last under strict scoring, because the judge reclassifies 93.9\%
of its strict errors as correct. It is tempting to attribute reclassification to output verbosity;
we test that directly rather than assume it, and it does not survive adjustment for task and model
composition, a negative result we report as such.

Item-level analysis compounds the finding. 148 of the 406 items (36.5\%) are answered identically by
all twelve models and separate no pair of systems; a predefined higher-discrimination band captures
134 items. Nominal size and discriminative coverage are not the same quantity, and benchmarks rarely
report the difference.

We also anchor the scale against a person. On a 60-item subset drawn from the medium and hard
strata, one non-specialist evaluator scored 80.0\% (48/60; 95\% Wilson CI [68.2, 88.2]). This is a
single evaluator and a reference point, not a population estimate: paired bootstrap tests against
all twelve models find no significant difference in either direction once corrected for the
twelve-test family. Dataset, evaluation harness, per-item judge verdicts,
and all model predictions are released.
\end{abstract}


\section{Introduction}

A benchmark leaderboard reports one number per model, and the field reads that number as a property
of the model. It is not. Accuracy is a joint product of what the model produced and what the scoring
rule was willing to accept, and for open-ended tasks those two things come apart. A model that
computes the right dividend and writes it as \texttt{19,000 crore} where the reference says
\texttt{Rs.\ 19,000 crore} has reasoned correctly and scored zero. Under string-matching evaluation
this is indistinguishable from not knowing the answer.

That string-match and judge-based scoring can disagree is not itself a new observation. What this paper
adds is a measurement of how large and structured that disagreement is under controlled
conditions---the same 406 items, the same twelve models, a domain chosen for regulatory text that is
less represented in common Western financial benchmarks---rather than an isolated anecdote: an
eleven-rank reversal for one model, a field-wide
compression to a third of its apparent spread, and a separate item-level analysis showing that
roughly half the items carrying that leaderboard don't discriminate between the systems on it at all.

The usual response is to treat such cases as noise---a small, uniform tax on every model that leaves
rankings intact. We show that this assumption fails: strict and judge-based scoring diverge sharply
for some models and barely at all for others. It is tempting to attribute this to output verbosity,
and we tested that explanation directly rather than asserting it: once task type and model identity
are accounted for, response length does not predict which strict failures a judge reclassifies
(Section~\ref{sec:regime}). The disagreement is real and large enough to reorder the leaderboard; what
drives it item by item is not yet established, and we report that as an open question rather than
papering over it with an intuitive but untested mechanism.

Testing this benefits from a domain where scoring effects are less likely to be confounded with
memorised recall of common benchmark text. Indian financial regulation is a reasonable candidate.
SEBI and RBI instruments embed numerical thresholds directly in prose, chain amendments across
decades so that resolving a question requires tracking which circular superseded which, and use
jurisdiction-specific terminology (LODR, PMLA, SFB, AIF, FEMA) that models trained predominantly on
US and European financial corpora have comparatively less reason to have encountered as verbatim
training text. We cannot verify what any evaluated model's training data contained, so this domain
choice reduces rather than rules out a familiarity confound. Existing financial NLP benchmarks are
built almost entirely from US or European sources (Section~\ref{sec:related}), so this setting is
also unmeasured on its own terms.

We introduce \textbf{IndiaFinBench}\footnote{Dataset, evaluation harness, and all model
predictions: \url{https://github.com/Rajveer-code/IndiaFinBench}.}, 406 expert-annotated items drawn from 34 of 192 collected SEBI and RBI
documents spanning 1992--2026, across four task types: regulatory interpretation, numerical
reasoning, contradiction detection, and temporal reasoning. Annotation quality is validated by a
model-based secondary pass (90.7\% agreement, 150 items) and an independent human
inter-annotator study (180 items, 44.3\% of the benchmark; 86.1\% overall agreement,
$\kappa = 0.712$ on the binary contradiction task). The benchmark is the instrument for the
measurement question, and we release it so the measurement can be repeated elsewhere.

Evaluating twelve contemporary LLMs zero-shot under a full-coverage cross-model judge audit
(Section~\ref{sec:judge})---every REG/NUM/TMP prediction, not only the ones strict scoring
flagged---produces the paper's central result. Scoring the same predictions under strict string
matching (75.12--89.66\%) and under the judge's own verdict (87.19--94.83\%) shows no positive rank
correspondence: Spearman $\rho = -0.273$ ($p = 0.39$, not significant) across the twelve models, and only
one of twelve (Kimi K2, rank 5 under each) holds its rank under both. A third, operational regime---strict-or-judge, accepting either
verdict---compresses the field further (94.1--98.0\%) and produces a third ordering again; under it,
DeepSeek-R1-Distill-Llama-70B (hereafter DeepSeek-R1-Distill), 12th of twelve under strict scoring,
ties for 1st, because 93.9\% of its strict errors are reclassified as correct by the cross-model
judge. That reversal illustrates how far a single asymmetric scoring choice can move a model, but
the paper's core claim does not rest on it: strict scoring and the judge's own verdict already
disagree enough, on their own, that neither is a reliable guide to which model reasons
better---though strict scoring remains the right measurement if what you care about is deployment
behaviour under a format-sensitive pipeline. Those are different questions, and accuracy alone does
not distinguish them.

A second measurement problem sits underneath the first. Counting items is not the same as counting
evidence: 148 of the 406 items (36.5\%) are answered identically by every model and separate no
pair of systems, while a predefined higher-discrimination band captures 134 items carrying
substantially more of the benchmark's discriminative signal. We report this for our own benchmark
because it is the kind of quantity benchmarks generally do not report, and because discriminative
coverage, not nominal size, determines how much a leaderboard difference can be trusted.

Our contributions, in the order the paper reports them: (1) a comparison of twelve LLMs under two
primary scoring regimes and one operational composite, showing the primary regimes show no positive
rank correspondence and that no model holds its rank across all three; (2) a direct test, not
an assumption, of whether output length explains which strict failures a judge reclassifies --- it
does not survive adjustment for task and model composition, a negative result we report as such
rather than the intuitive but untested mechanism it replaces; (3) a discriminative-coverage analysis
using an exact pairwise-disagreement measure, showing 148 items separate no pair of systems while
134 carry most of the benchmark's discriminative signal; (4) IndiaFinBench
itself, with a validated annotation protocol and published agreement statistics; (5) a human
reference point reported with its limits stated; and (6) full public release of the dataset, harness,
per-item judge verdicts, and all model predictions.


\section{Related Work}
\label{sec:related}

\paragraph{Financial and regulatory benchmarks.}
The financial NLP community has built substantial evaluation infrastructure, anchored almost
entirely in Western text. FinQA \citep{chen2021finqa} and ConvFinQA \citep{zheng2022convfinqa} test
numerical reasoning over SEC filings; FiNER-139 \citep{loukas2022finer} targets entity recognition
in the same filings; FLUE \citep{shah2022flue} packages several such tasks into a multi-task suite.
PIXIU/FLARE \citep{xie2023pixiu} and FinBen \citep{xie2024finben} push toward breadth, the latter
spanning 42 datasets across 24 tasks. FinanceBench \citep{islam2023financebench} is closest to our
own construction---expert-verified gold answers over financial documents---but covers publicly
listed US companies only. Legal NLP contributes adjacent structure without the financial content:
CUAD \citep{hendrycks2021cuad} for contract clauses, LexGLUE \citep{chalkidis2022lexglue} for
European legal text, ILDC \citep{malik2021ildc} for Indian court judgments. Across this literature
the binding constraint is geography rather than task diversity, and no benchmark we are aware of
defines amendment-chain supersession as a task type.

\paragraph{How benchmarks are scored.}
The methodological gap matters more to this paper. Extractive QA evaluation inherits its scoring
conventions from SQuAD \citep{rajpurkar2016squad} and DROP \citep{dua2019drop}: exact match and
token-overlap F1 against a reference string. Those conventions were designed for short spans copied
from a passage, where the space of correct surface forms is small. Applied to open-ended
quantitative answers---where \texttt{Rs.\ 19,000 crore}, \texttt{19,000 crore}, and
\texttt{nineteen thousand crore} are the same answer---they measure formatting as much as
correctness. On extractive QA specifically, exact match and F1 correlate weakly with human judgments
(0.22, 0.40) where LLM-as-judge scores correlate far more strongly, up to 0.85
\citep{ho2025reassessing}---the same gap this paper measures, on open-ended regulatory answers rather
than short spans. LLM-as-judge evaluation has emerged as the usual alternative
\citep{zheng2023judging,gu2024survey}, and its own failure modes are documented: position, verbosity,
and self-enhancement biases \citep{zheng2023judging}, familiarity and anchoring effects
\citep{wang2024fairevaluators}, a preference for a judge's own family's outputs
\citep{wataoka2024selfpreference}, and raw agreement rates that overstate discriminative reliability
once corrected for chance \citep{norman2026reliability}. We are not claiming to discover that judges
can be biased; that is established. Our contribution is measuring, on one benchmark, how large the disagreement between a
string-matching pipeline and a cross-model judge actually is, in both directions, and how much that
disagreement alone---independent of which specific bias produces it---can move a twelve-model
leaderboard. Our position is not that one regime is correct. It is that the two regimes answer
different questions, that the difference is large enough to reorder a leaderboard, and that reporting
either one alone leaves the reader unable to tell capability from format compliance. Holistic
frameworks such as HELM \citep{liang2022helm} argue for multi-metric reporting on related grounds;
we supply a concrete case where the metric choice, not the metric count, is what moves the ranking.

\paragraph{What a benchmark reports about itself.}
Benchmarks are commonly described by nominal item count. Whether those items discriminate between
the systems being ranked is rarely stated, though the psychometrics literature has long treated item
difficulty and discrimination as basic descriptive properties of a test. Section~\ref{sec:effsize}
reports them for IndiaFinBench.


\section{Dataset Construction}

Closing this gap requires building from Indian regulatory documents at a scale sufficient for
reliable model comparison, while preserving the domain expertise that interpreting them
demands---the constraint that shaped every design choice below.

\subsection{Source Document Collection}

We collected 192 regulatory documents spanning 1992--2026 from two official Indian government
sources: the Securities and Exchange Board of India (92 documents: circulars, master circulars,
regulations, orders; sebi.gov.in) and the Reserve Bank of India (100 documents: circulars, policy
statements, directions; rbi.org.in). Documents were downloaded using a custom Python scraping
pipeline and converted to clean text using pdfplumber, which handles the multi-column layouts and
embedded tables common in Indian regulatory PDFs particularly well. Figure~\ref{fig:pipeline}
(Appendix~\ref{app:examples}) illustrates the full dataset construction pipeline;
Appendix~\ref{app:sourcedocs} lists the regulatory categories covered.

The 406 annotated items are drawn from \textbf{34 of these 192 documents}---the collection was
sized for coverage of the regulatory categories in Appendix~\ref{app:sourcedocs}, while annotation
effort concentrated on documents dense enough in extractable QA content to be worth an expert
annotator's time. Section~\ref{app:provenance} reports the full provenance: item counts per
document (median 6, largest 63) and the document-level clustering used for the significance testing
in Section~\ref{sec:significance}, since items sharing a source document are not independent draws.

\subsection{Task Types}

IndiaFinBench defines four task types, each probing a distinct reasoning capability
(Table~\ref{tab:tasktypes}); Figure~\ref{fig:taskexamples} (Appendix~\ref{app:examples}) gives a
full worked example of each.

\begin{table}[t]
\centering\small
\begin{tabular}{lrp{9.7cm}}
\toprule
\textbf{Task} & \textbf{N} & \textbf{Definition} \\
\midrule
Regulatory Interpretation (REG) & 174 & Identify the correct rule, compliance threshold, or scope
of applicability from a regulatory passage---e.g., that a stock exchange must forward a registration
application ``not later than thirty days of receipt.'' \\
Numerical Reasoning (NUM) & 92 & Perform arithmetic over figures embedded in regulatory
text---e.g., a Small Finance Bank's maximum eligible dividend from its Tier 1 Capital Ratio and
adjusted profit after tax---requiring correct extraction \emph{and} execution. \\
Contradiction Detection (CON) & 62 & Given two passages from different regulatory instruments,
determine (Yes/No, one-sentence explanation) whether they contradict, probing supersession
tracking across frequently-amended circulars. \\
Temporal Reasoning (TMP) & 78 & Establish chronological ordering of regulatory events, identify
which rule version was in force at a given time, or compute elapsed time between milestones
referenced only by date. \\
\bottomrule
\end{tabular}
\caption{IndiaFinBench task types.}
\label{tab:tasktypes}
\end{table}

\subsection{Annotation Protocol}

All question-answer pairs were authored by the primary annotator, who has specialist experience with
Indian financial regulatory documents spanning SEBI's securities regulation framework, RBI's
prudential norms, and the amendment-chain structure linking hundreds of circulars over three
decades---expertise unavailable to crowdsourced pools, so expert single-annotator construction is
standard for domain-specific benchmarks \citep{hendrycks2021cuad, malik2021ildc,
islam2023financebench}, validated empirically by an independent human inter-annotator agreement
study (Section~\ref{sec:iaa}). Each item was individually reviewed for an unambiguous, unique answer
derivable from context alone, and consists of a context passage (80--500 words), a question, a
reference answer, and metadata fields (task type, difficulty, source document).

Answer formats are standardized by task type: extractive spans for regulatory interpretation and
temporal reasoning; calculated values with units for numerical reasoning (e.g., `Rs.~5,500 crore');
and `Yes' or `No' with a brief explanation for contradiction detection.

Difficulty levels were assigned by the number of reasoning steps required: easy items are
single-step extraction (160 items, 39.4\%), medium items require multi-clause reasoning or
calculation (182, 44.8\%), and hard items span multiple instruments or complex arithmetic (64,
15.8\%; Section~\ref{sec:difficulty} finds this label predicts empirical difficulty for only three
of the twelve evaluated models). The 406-item scale reflects a deliberate quality-over-quantity
decision, structurally incompatible with crowdsourcing (cf.\ FinanceBench's 150 expert-verified
items, Section~\ref{sec:related}).

\subsection{Model-Based Secondary Validation}

To confirm that items are unambiguously answerable from context, a secondary validation pass used
LLaMA-3.3-70B (via Groq API) as a secondary model-based quality check under a context-only, zero-shot
prompt (temperature = 0), accessed in isolation from the evaluation pipeline. Although LLaMA-3.3-70B
also appears in the main evaluation, the two uses are functionally distinct: this pass asks whether a
question is unambiguously answerable from its context passage, a different task from the
evaluation's open-ended QA, following established benchmark construction practice
\citep{islam2023financebench, hendrycks2021cuad}.

\begin{table}[t]
\centering\small
\begin{tabular}{lrrrr}
\toprule
 & \multicolumn{2}{c}{\textbf{Model-based}} & \multicolumn{2}{c}{\textbf{Human IAA}} \\
\textbf{Task} & \textbf{Items} & \textbf{Agr.} & \textbf{Items} & \textbf{Agr.} \\
\midrule
REG & 53 & 85.7\% & 63 & 92.1\% \\
NUM & 32 & 84.4\% & 44 & 81.8\% \\
CON & 30 & 96.7\%\textsuperscript{$\kappa$=.918} & 35 & 80.0\%\textsuperscript{$\kappa$=.712} \\
TMP & 35 & 77.1\% & 38 & 86.8\% \\
\midrule
\textbf{All} & \textbf{150} & \textbf{90.7\%} & \textbf{180} & \textbf{86.1\%} \\
\bottomrule
\end{tabular}
\caption{Annotation validation: model-based secondary pass (150-item subset) and human
inter-annotator agreement (180 items, three 60-item rounds, second annotator blind to reference
answers). Cohen's $\kappa$ applies only to the binary CON task.}
\label{tab:validation}
\end{table}

The secondary pass agreed with the primary annotation on 90.7\% of the sampled items
(Table~\ref{tab:validation}, left); items with genuine disagreement ($\sim$1.3\% of the initial set)
were removed.

\subsection{Human Inter-Annotator Agreement}
\label{sec:iaa}

To provide rigorous empirical validation of single-annotator annotation quality, we conducted a
human inter-annotator agreement (IAA) study in which a second independent human annotator completed
three annotation rounds of 60 items each (180 items total; 44.3\% of the benchmark), drawn
proportionally from all four task types and weighted toward medium and hard difficulty. The
annotator worked from context passages and questions alone, without access to the primary
annotator's reference answers at any stage.

Agreement was then computed between the primary annotator's reference answers and the second
annotator's responses, using the same four-stage scoring procedure applied to model predictions
(Section~\ref{sec:scoring}); results are in Table~\ref{tab:validation} (right).

The $\kappa$ = 0.712 for contradiction detection falls in the `substantial agreement' band by
\citet{landiskoch1977} conventions, and is comparable to human agreement rates reported for similar
binary contradiction detection tasks in legal NLP. Numerical reasoning (81.8\%) and contradiction
detection (80.0\%) are the two lowest task-level agreement rates, both driven by the same mechanism:
reference answers are often terse final values, while the second annotator's answers frequently embed
a full derivation or restate the question context around the same number, so the strict scorer's
exact and numeric-set matching stages can miss an answer that is substantively identical but longer.
We manually reviewed and independently re-verified the arithmetic or logic of all eight discordant
numerical-reasoning items across the three rounds and recorded the review, item by item, in
\texttt{annotation/iaa/num\_discordance\_review.csv} (released with the benchmark): every case's
computed values or final conclusions were equivalent; disagreements arose solely from intermediate-step
presence, currency symbol style, unit labelling, date format, or reasoning-path verbosity, never from a
numeric or substantive-conclusion discrepancy.

\section{Experimental Setup}

\subsection{Models}
\label{sec:models}

We evaluate twelve models spanning a wide range of sizes, providers, and access modes on the full
406-item benchmark: Gemini 2.5 Flash, Gemini 2.5 Pro, Qwen3-32B, LLaMA-3.3-70B, Llama 4 Scout 17B,
Kimi K2, GPT-OSS 120B, GPT-OSS 20B, DeepSeek-R1-Distill-Llama-70B, LLaMA-3-8B, Mistral-7B, and
Gemma 3 4B (the last three run locally via Ollama v0.6.5, 4-bit quantization, on an RTX 4060,
8~GB; exact providers, API identifiers, and completion budgets: Table~\ref{tab:models},
Appendix~\ref{app:models}). All models were evaluated zero-shot with no fine-tuning or prompt
adaptation, at temperature 0.0 --- but \textbf{not} under a shared completion budget: the harness set
per-provider generation limits (200--2048 tokens) for cost reasons unrelated to any experimental
design choice, and Gemini 2.5 Pro received no explicit cap at all. Budget heterogeneity is a
plausible confound for any format-sensitivity claim this paper makes, and we treat it as one rather
than assuming it away. Appendix~\ref{app:models} reports the exact provider, API model identifier or
Ollama tag, and completion budget used for every model, drawn directly from the evaluation harness
rather than restated from memory; Appendix~\ref{app:truncation} separately quantifies how much of the
logged prediction text this budget heterogeneity, and the harness's own logging limits, actually
truncate, and reports the paper's headline result restricted to the truncation-free item subset.
Three additional models --- Claude 3 Haiku (150 items), NVIDIA Nemotron 3 Super 120B (140 items),
and Qwen3-235B (120 items) --- were evaluated on partial subsets during earlier exploratory passes
and are excluded from every result in this paper because they were never run on the full 406-item
benchmark; their raw predictions remain in the public release for transparency, unused for any
reported number.

\subsection{Prompting Strategy}

Our evaluation follows the extractive QA paradigm established by SQuAD \citep{rajpurkar2016squad},
in the same spirit as DROP's discrete-reasoning-over-text framing \citep{dua2019drop}, with context
passages provided directly to the model under zero-shot, context-only constraints. This isolates
the model's ability to reason about the supplied regulatory text from its ability to recall
memorized facts; it does not by itself establish that the underlying regulatory documents are
absent from any model's training corpus, only that the model cannot use such memorized content
\emph{instead of} the supplied passage without the context-only instruction being violated. All
models received a system prompt establishing the context-only constraint (full text:
Appendix~\ref{app:prompt}). Task-specific user prompts provided appropriate formatting instructions
for each task type. For
contradiction detection, both passages were labelled explicitly as `Passage A / Passage B'. For
numerical reasoning, models were instructed to show calculation steps and include units---an
instruction in tension with the system prompt's ``no preamble'' constraint, which is one source of
the format-versus-content gap that Section~\ref{sec:regime} measures directly. All models received
the identical prompts described above at temperature 0.0; they did \textbf{not} share a completion
budget (Appendix~\ref{app:models} reports each model's exact value), which is a separate,
independently disclosed source of output-length variation from the prompting tension just described.

We evaluate exclusively under zero-shot conditions, as this most closely reflects practical
deployment where users query models without domain-specific priming, and because it eliminates
confounds from example selection strategy and ordering effects. Few-shot results for four top
models are reported in Appendix~\ref{app:fewshot}.

\subsection{Scoring}
\label{sec:scoring}

Answers were scored using a multi-stage matching procedure applied in sequence: (1) exact match
after case-normalization and punctuation stripping; (2) fuzzy token match using RapidFuzz
\texttt{token\_set\_ratio} $\geq$ 0.72, applied when exact match fails, with the 0.72 threshold
established by manual inspection of 20 borderline cases and validated against adjacent thresholds
(0.65 and 0.80); (3) numerical extraction match, where the item is scored correct if the set of
numbers extracted from both the reference and prediction are identical (handling currency symbols,
comma separators, and units); and (4) Yes/No match for contradiction detection, comparing the
leading word of the prediction to the reference. We refer to accuracy under this procedure as
\textbf{strict} accuracy throughout.

Section~\ref{sec:regime} evaluates the same predictions under two further regimes, built from a
cross-model judge (Section~\ref{sec:judge}) that reviews every prediction on REG, NUM, and TMP---not
only the ones strict scoring marks wrong: \textbf{judge-only}, in which the judge's verdict is final
in both directions, and \textbf{judge-augmented}, the asymmetric composite strict-OR-judge (a
strict-correct answer counts even if the judge disputes it; a strict-wrong answer counts if the judge
accepts it). Judge-only and strict are the paper's two primary regimes; judge-augmented is an
operational sensitivity regime, included to quantify what happens when a judge is applied only to
overturn strict failures and never to reject strict-correct ones, not because we consider it
authoritative. We avoid ``corrected,'' ``semantic,'' or unqualified ``judge-audited'' throughout:
the judge is itself an imperfect measurement (Section~\ref{sec:judge} reports its own error rate
against human adjudication), not a ground truth that strict scoring falls short of, and each of the
three regimes needs its own name because they can and do disagree.

DeepSeek-R1-Distill-Llama-70B was accessed via OpenRouter through an endpoint that returns final
answers without exposed reasoning traces; its stored predictions contain no chain-of-thought markup,
so we make no claim about the content of its internal deliberation, only about its final output
under its own completion budget (2048 tokens, the largest of any model in the study --- see
Appendix~\ref{app:models}), which rules out output-budget truncation as an explanation for its
strict-scoring behaviour specifically.


\section{The Scoring Regime Reorders the Leaderboard}
\label{sec:regime}
\label{sec:budget}

Every model was scored three times over identical predictions: \textbf{strict}, \textbf{judge-only},
and \textbf{judge-augmented}, all defined in Section~\ref{sec:scoring}. Strict and judge-only are
this paper's two primary regimes; judge-augmented is an operational sensitivity regime only, applying
a judge asymmetrically the way a practitioner who rescues strict failures but never challenges strict
successes effectively would.

Nothing about the models changed between regimes. Only the rule changed.

\begin{figure}[t]
\centering
\includegraphics[width=0.95\linewidth]{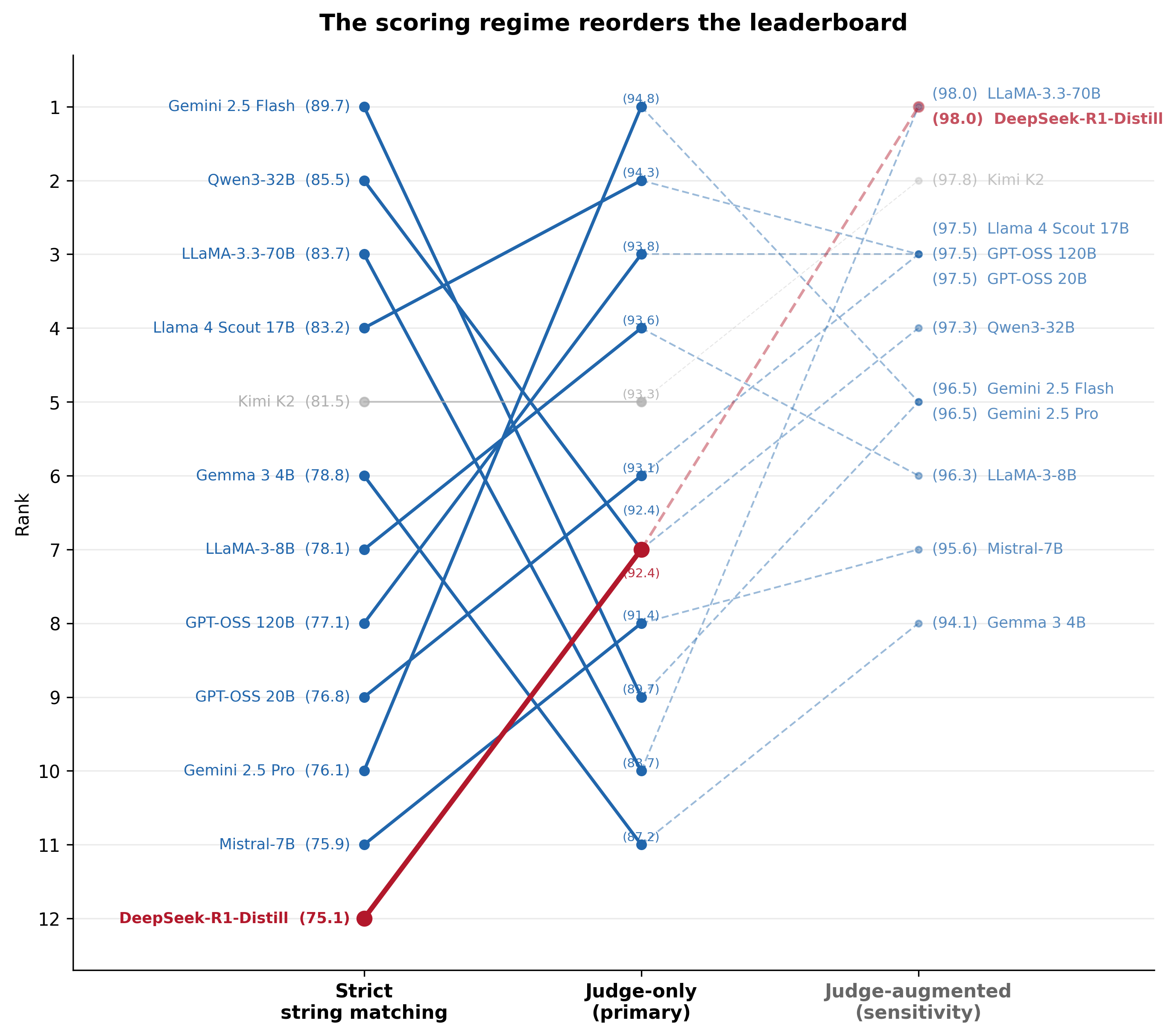}
\caption{Model rank under strict string matching (left), the judge's own verdict (centre), and the
judge-augmented composite (right), over identical predictions. The centre column is this paper's
second primary regime; the right column is a sensitivity regime, shown in a lighter tone. Coloured
lines move by two or more positions between strict and judge-only. Tied scores share a rank.}
\label{fig:regime}
\end{figure}

Appendix~\ref{app:regimetable} tabulates exact accuracy and rank for all twelve models under all
three regimes; the discussion below states the comparisons that matter.

\textbf{Strict and judge-only show no positive rank correspondence.} Strict accuracy spans
75.12--89.66\%; judge-only accuracy spans 87.19--94.83\%, with a different leader (Gemini 2.5 Pro)
and a different trailer (Gemma 3 4B) than strict scoring's (Gemini 2.5 Flash, DeepSeek-R1-Distill).
Across the twelve models, Spearman $\rho = -0.273$ ($p = 0.39$), Kendall $\tau = -0.168$ ($p = 0.45$).
With twelve models these tests have little power, so we do not claim the rankings are negatively
correlated in a strict statistical sense---we claim they show no positive correspondence, which
Figure~\ref{fig:regime} and Appendix~\ref{app:regimetable} show directly: only one of twelve models
(Kimi K2, rank 5 under each) holds the same rank under both. Neither regime is more correct than the other; each answers a different question
(exact reproduction of a reference format, versus a second model's judgment of substantive
correctness), and a benchmark that reports only one has silently chosen which question to answer.

\textbf{The judge-augmented sensitivity regime illustrates how far a single asymmetric choice can
move a model.} Under strict-OR-judge, DeepSeek-R1-Distill---12th of twelve, last place, under
strict scoring---reaches an exact tie for 1st with LLaMA-3.3-70B (both 398/406, 98.0\%). Of its 99
strict errors on REG/NUM/TMP, 93 are reclassified as correct by the cross-model judge---93.9\%---
leaving 6 residual failures, the fewest of any model in the study. Under the judge's own verdict
alone (judge-only, not augmented), DeepSeek-R1-Distill ranks 7th of twelve, tied with Qwen3-32B
(92.4\%): the judge
recovers most of its strict failures but also disputes some of its strict successes, and the
augmented regime's asymmetric rule---rescue but never reject---is what produces the 1st-place tie.
We report this as what it is: a real, measured consequence of one specific, defensible-but-partial
way of combining two scores, not evidence that DeepSeek-R1-Distill's regulatory reasoning is
best-in-field. Its strict score is a measurement under a particular operational
definition---exact and near-exact string match against a reference format---not a direct measurement
of its regulatory reasoning; under that definition its output sits far from the reference, and the
model narrates its working where the system prompt asked for an answer with no preamble. We tested,
rather than assumed, whether output length itself explains which strict failures get reclassified
(Section~\ref{sec:discussion}); it does not survive adjustment for task and model composition, so we
do not claim narration or verbosity as the mechanism---only that the judge and strict scoring
disagree on this model's output more than on any other's, for reasons this paper's evidence does not
fully resolve.

A second effect is visible in the accuracies rather than the ranks. The full field spans 14.5 points
under strict scoring (75.12--89.66) and just 3.9 points under judge-augmented scoring (94.1--98.0);
judge-only compresses it to 7.6 points (87.19--94.83). Compression under either judge-involving
regime is real, but the size of the effect depends on which one: judge-augmented's asymmetric rule
compresses roughly twice as much as judge-only's symmetric one, because it can only ever move scores
upward.

This has a practical reading that is not a criticism of string matching. A compliance workflow that
consumes model output through a format-sensitive parser really would fail on DeepSeek-R1-Distill's
93 reclassified items, and for that deployment question the strict number is the honest one. What
the strict number cannot isolate is underlying regulatory correctness from reference-format
compliance, and the two are routinely reported as if they were the same quantity.

\subsection{Evaluation-Configuration Sensitivity}
\label{sec:budget-results}

Completion budgets and write-time logging caps were not uniform across the original evaluation
(Section~\ref{sec:models}). We re-ran 11 of 12 models at a matched, uncapped 512-token budget under
the same four-stage scorer and items (Appendix~\ref{app:matched-budget-results}; Gemini 2.5 Pro is
permanently unreachable on any fresh API key). Ten of the eleven models shifted by at most 4.2
points in either direction, consistent with a modest, mixed-sign operational effect layered on top
of a change of transport for five of them; DeepSeek-R1-Distill is an outlier at $+10.59$pp, reported
without endorsing a single explanation (Appendix~\ref{app:budget-outlier} gives two live candidates
and the evidence against attributing it cleanly to either). What the re-run supports without
qualification is the general point: strict accuracy is sensitive to the combined evaluation
configuration---completion budget, logging configuration, and serving provider jointly, not any one
in isolation---sometimes by several points. Further evidence, alongside the scoring-regime result,
that a single leaderboard number is a property of the measurement protocol as much as of the model.

\section{Discriminative Coverage: How Many Items Actually Discriminate}
\label{sec:effsize}

A benchmark's nominal size is the number of items it contains. Its \textbf{discriminative
coverage}, for the purpose of separating systems, is the number of items on which systems actually
differ. (We avoid ``effective size'': it invites confusion with effective-sample-size, a distinct
and more specific statistical quantity we do not compute.)

For an item on which $k$ of the twelve models answer correctly, the number of model \emph{pairs}
that item places on opposite sides of a correct/incorrect split is exactly $D(k) = k(12-k)$, out of
66 total pairs among twelve models. This is threshold-free and exact: $D(k)=0$ only at the two extremes
($k=0$ or $k=12$, every model agrees), rises to a maximum $D(6)=36$ at the most evenly split items,
and is symmetric ($D(k)=D(12-k)$). An earlier draft of this analysis defined a ``ceiling'' band as
$p \ge 0.9$ (i.e.\ $k \in \{11,12\}$) and described it as carrying almost no discriminative
information. That conflates two different items: a $k=12$ item genuinely distinguishes no pair
($D=0$), but a $k=11$ item still separates the one dissenting model from the other eleven ---
$D(11)=11$, or $11/66 \approx 16.7\%$ of all pairs. The same error affected the symmetric ``floor''
band ($k \in \{0,1\}$). We correct this below using $D(k)$ directly rather than a $p$-threshold.

\begin{table}[t]
\centering
\caption{Items by exact pairwise discrimination $D(k)=k(12-k)$, $k$=models answering correctly out
of 12. Zero-information items agree unanimously; every other item distinguishes at least 11 of the
66 model pairs. The predefined band $0.3<p<0.8$ ($k\in\{4,\ldots,9\}$) from Table~\ref{tab:matchedbudget}'s
companion analysis is retained as one descriptive slice, not as a claim that items outside it carry
no signal.}
\label{tab:effsize}
\begin{tabular}{lrrr}
\toprule
Category & Items & \% & $D(k)$ range \\
\midrule
Zero-information ($k=0$ or $12$)         & 148 & 36.5 & 0 \\
Near-unanimous ($k=1$ or $11$)           & 72  & 17.7 & 11 \\
Predefined band ($0.3<p<0.8$, $k\in\{4..9\}$) & 134 & 33.0 & 27--36 \\
Remaining ($k\in\{2,3,10\}$)             & 52  & 12.8 & 20--27 \\
\midrule
Total                                     & 406 & 100  & \\
\bottomrule
\end{tabular}
\end{table}

143 items are answered correctly by all twelve models and 5 by none; these 148 (36.5\%) are the
benchmark's true zero-information items under any scoring rule, since no pair of systems is
separated by them. A further 72 items (17.7\%) sit at $k=11$ or $k=1$ --- near-unanimous, but each
still distinguishes 11 of 66 pairs, so we do not group them with the zero-information set. Averaged
over all 406 items, the mean pairwise-disagreement fraction is $\overline{D}/66 = 0.229$: on a
uniformly random pair of the twelve models, an average item has a 22.9\% chance of ranking them
differently. The predefined band captures disproportionately more of this signal than its item
count alone suggests: its 134 items (33.0\% of the benchmark) carry 68.9\% of the benchmark's total
pairwise-discriminative mass ($\sum_i D(k_i)$ over all 406 items) --- not merely "most," but over
two-thirds of the evidence from one-third of the items.

We report this about our own benchmark deliberately. It bears directly on how the results above
should be read: with 148 items (36.5\%) contributing no pairwise information at all, a gap of a few
points between two mid-table models is supported by a smaller item pool than the nominal 406. It
also suggests a construction lesson we did not anticipate when building the benchmark---annotating
for coverage of a domain and annotating for discrimination between systems are different
objectives, and expert effort spent on items every competent model answers correctly buys
descriptive completeness rather than measurement power.

We do not claim this ratio is unusual. We claim it is unmeasured: we are not aware of comparable
benchmarks in this literature that report what fraction of their items separate the systems they
rank, and without that figure a reported accuracy gap cannot be sized against the evidence
supporting it.


\subsection{Overall Performance}
\label{sec:main-results}
\label{sec:main-results-text}

\begin{table*}[t]
\centering\small
\begin{tabular}{lrrrrrl}
\toprule
\textbf{Model} & \textbf{REG} & \textbf{NUM} & \textbf{CON} & \textbf{TMP} & \textbf{Overall} & \textbf{95\% CI} \\
\midrule
Gemini 2.5 Flash & 93.1 & 84.8 & 88.7 & 88.5 & 89.7 & [86.3\%, 92.3\%] \\
Qwen3-32B & 85.1 & 77.2 & 90.3 & 92.3 & 85.5 & [81.7\%, 88.6\%] \\
LLaMA-3.3-70B & 86.2 & 75.0 & 95.2 & 79.5 & 83.7 & [79.8\%, 87.0\%] \\
Llama 4 Scout 17B & 86.2 & 66.3 & 98.4 & 84.6 & 83.3 & [79.3\%, 86.6\%] \\
Kimi K2 & 89.1 & 65.2 & 91.9 & 75.6 & 81.5 & [77.5\%, 85.0\%] \\
Gemma 3 4B & 86.2 & 70.7 & 79.0 & 71.8 & 78.8 & [74.6\%, 82.5\%] \\
LLaMA-3-8B & 79.9 & 64.1 & 93.5 & 78.2 & 78.1 & [73.8\%, 81.8\%] \\
GPT-OSS 120B & 79.9 & 59.8 & 95.2 & 76.9 & 77.1 & [72.8\%, 80.9\%] \\
GPT-OSS 20B & 79.9 & 58.7 & 95.2 & 76.9 & 76.8 & [72.5\%, 80.7\%] \\
Gemini 2.5 Pro\textdagger{} & 89.7 & 48.9 & 93.5 & 64.1 & 76.1 & [71.7\%, 80.0\%] \\
Mistral-7B & 79.9 & 66.3 & 80.6 & 74.4 & 75.9 & [71.5\%, 79.8\%] \\
DeepSeek-R1-Distill & 72.4 & 69.6 & 96.8 & 70.5 & 75.1 & [70.7\%, 79.1\%] \\
\midrule
\textbf{Average} & \textbf{84.0} & \textbf{67.2} & \textbf{91.5} & \textbf{77.8} & \textbf{80.1} & --- \\
\bottomrule
\end{tabular}
\caption{IndiaFinBench results --- accuracy (\%) by task type under the strict string-matching
pipeline. All twelve models evaluated on the full 406-item benchmark. 95\% Wilson score
\citep{wilson1927} confidence intervals shown for overall accuracy; these are descriptive item-level
intervals, and pairwise significance is assessed separately with the document-clustered bootstrap
robustness analysis (Appendix~\ref{app:provenance}). \textdagger{}Gemini 2.5 Pro
evaluated via Vertex AI; its lower NUM/TMP scores narrow sharply under either judge-involving
regime (Section~\ref{sec:regime}), a scoring-rule effect rather than a measured verbosity
mechanism. Judge-only and judge-augmented accuracies are in Section~\ref{sec:regime} and
Appendix~\ref{app:judge}.}
\label{tab:mainresults}
\end{table*}

Table~\ref{tab:mainresults} reports strict accuracy for all twelve models on the full 406-item
benchmark. Scores span 75.12\% (DeepSeek-R1-Distill, 305/406) to 89.66\% (Gemini 2.5 Flash, 364/406),
a 14.5-point (59-item) range
that is wide enough to separate the field but narrow enough that most adjacent pairs are not
individually distinguishable.

The task means are more informative than the model means. Averaged across models, contradiction
detection is highest at 91.5\%, followed by regulatory interpretation at 84.0\%, temporal reasoning
at 77.8\%, and numerical reasoning at 67.2\%. That ordering is stable: Kendall's $W$ across the four
task profiles is 0.853 ($\chi^2 = 30.7$, $p < 0.001$, $n = 12$), so the models largely agree about
which tasks are hard, even where they disagree about their own overall standing. Difficulty here is
a property of the tasks rather than of any particular system.

Two of these numbers should not be read at face value, for opposite reasons. The contradiction
figure is inflated by class imbalance and is addressed in Section~\ref{sec:tasklevel}. The Gemini
2.5 Pro row is depressed by scoring: at 76.1\% strict it sits tenth, below the much smaller Gemini
2.5 Flash, and the gap narrows sharply under either judge-involving regime---to 96.5\% judge-augmented,
94.8\% judge-only (the highest of any model under judge-only). It is the same strict-versus-judge disagreement documented at length in
Section~\ref{sec:regime}, visible here as a single anomalous row.

\subsection{Task-Level Analysis}
\label{sec:tasklevel}

The four tasks differ sharply in how well they separate systems.

\textbf{Numerical reasoning discriminates most.} Its 35.9-point spread (84.8\% down to 48.9\%) is
the widest of any task. Multi-step arithmetic over thresholds embedded in regulatory prose requires
correct extraction \emph{and} correct execution, and failure at either stage produces a wrong
answer.

\textbf{Contradiction detection discriminates least, despite the highest mean.} 53 of its 62 items
carry the gold label `No', so a classifier that always answers `No' scores 85.5\%. Against that
baseline the 91.5\% task mean is a 6.0-point margin, not a strong result, and 10 of 12 models
exceed the baseline while Mistral-7B (80.6\%) and Gemma 3 4B (79.0\%) fall below a trivial
majority classifier. High accuracy on an imbalanced binary task is close to uninformative, and we
report balanced accuracy in Appendix~\ref{app:con} rather than leaning on the raw figure.

\textbf{Regulatory interpretation and temporal reasoning sit in between}, at 20.7-point and
28.2-point spreads. Temporal reasoning is where the top of the table separates: Qwen3-32B leads at
92.3\% while Gemini 2.5 Pro reaches only 64.1\%.

\subsection{Difficulty Stratification}
\label{sec:difficulty}

Items were labelled easy (160), medium (182), or hard (64) at annotation time by the number of
reasoning steps required. If that label tracked model difficulty as intended, accuracy should fall
monotonically from easy to hard. It does, for exactly three of twelve models, most sharply Gemini 2.5
Pro (83.1\% $\to$ 68.8\%, a 14.4-point decline); the other nine do not decline monotonically at
all---the dominant pattern in this benchmark is not the one the difficulty label was designed to
produce (full breakdown, trajectories, and the most striking individual case:
Appendix~\ref{app:difficulty}). We read this as evidence about the difficulty label, not about any
individual model, and flag it as a limitation of the annotation scheme
(Section~\ref{sec:limitations}) rather than adjusting the labels post hoc. It strengthens the case
for the discriminative-coverage analysis in Section~\ref{sec:effsize}: nominal difficulty and
empirical difficulty are visibly different quantities on this benchmark, in the same way nominal
size and discriminative coverage are.

\subsection{A Human Reference Point}
\label{sec:human}

To anchor the accuracy scale, a single non-specialist evaluator---a university-educated adult with
no professional background in finance or law---answered a 60-item subset drawn from the medium and
hard strata, under the same context-only constraint given to the models and scored by the identical
pipeline.

The evaluator scored 80.0\% (48/60; 95\% Wilson CI [68.2\%, 88.2\%]), with a task profile of
REG 100\% (11/11), TMP 87.5\% (14/16), CON 82.4\% (14/17), and NUM 56.2\% (9/16). On the same 60
items, five of the twelve models score above the evaluator and seven below. We ran a paired
bootstrap test against the evaluator for all twelve models, not only the extremes, since reporting
only the best and worst comparison raises its own multiplicity concern. One model, Gemma 3 4B at
66.7\%, is nominally significant at $p = 0.02$ uncorrected; none---including Gemma 3 4B---survives
Bonferroni correction across the twelve-test family ($\alpha = 0.05/12 \approx 0.0042$). The honest
reading is that this single-evaluator comparison cannot statistically distinguish any of the twelve
models from the human reader on this subset, in either direction; the best model, Gemini 2.5 Flash
at 85.0\% against the evaluator's 80.0\%, is furthest from significant ($p = 0.44$).

This comparison's limits are severe: one evaluator on 60 items supports no population-level claim,
and a domain specialist would almost certainly score higher, particularly on numerical items where a
finance professional would recognise calculation forms this evaluator had to derive from the
passage. It anchors the scale rather than estimating human capability---reported because the
alternative, no human referent at all, leaves the reader no way to judge whether 85\% on this
benchmark is good.

One asymmetry does survive those caveats and is worth recording. The evaluator was perfect on
regulatory interpretation and worst on numerical reasoning, which is the same ordering the models
produce. The hardest task for this benchmark's models is also the hardest for a careful human
reader, which suggests the difficulty is a property of multi-step quantitative inference over
regulatory prose rather than an artifact of how models are prompted or scored.


\subsection{Cross-Model Judge}
\label{sec:judge}

An LLM-as-judge audit is only as convincing as the judge's independence from what it is judging.
The judge used in our initial pilot (Appendix~\ref{app:judge}) was Gemini 2.5 Flash---one of the
twelve evaluated models. To assess whether this affects the audit, we additionally judge every REG/NUM/TMP
prediction with \textbf{phi4-mini} (Microsoft Phi, 3.8B parameters), run locally via Ollama at zero
cost. Phi shares no model family with any of the twelve benchmarked systems. We call this a
\textbf{cross-model} judge, not an independent or ground-truth one: phi4-mini is a small model, and
disagreement between it and Gemini 2.5 Flash is compatible with either judge being wrong, which is
why we pair it with human adjudication rather than treating its verdict as final.

The cross-model pass covers \textbf{every} REG/NUM/TMP prediction---344 items $\times$ 12 models
$=$ 4{,}128 judgements---not only the items strict scoring marks wrong. Full coverage is what makes
a \emph{judge rejection rate of strict-correct predictions} estimable for the first time: the
original audit design, by construction, could only characterise strict-to-judge overturns. CON is excluded from both judges, matching
the pipeline's existing rationale (Section~\ref{sec:scoring}): it is scored by exact Yes/No match
against an unambiguous binary label, which does not benefit from semantic review. Excluding it is
conservative for the disagreement rates we report below, not convenient: CON is the one task where
strict and judge-based scoring would agree almost by construction, so its absence removes a source
of trivial agreement rather than manufacturing disagreement elsewhere.

The judge rubric (Appendix~\ref{app:judge}) accepts numeric answers within 1\% rounding of the
reference; Appendix~\ref{app:judge-tolerance} tests that specific tolerance against two stricter
alternatives and finds the strict-versus-judge-only claim above holds at all three, though
individual models' judge-only ranks are more tolerance-sensitive than the headline spread suggests.

Before committing to the full pass we validated the judge on two small, targeted samples that
confirmed it both overturns genuine format mismatches and retains cases where the substantive
answer is wrong (Appendix~\ref{app:judge}).

Table~\ref{tab:judgereliability} reports, per model, the strict-to-judge overturn rate, the judge
rejection rate of strict-correct predictions (estimable for the first time here, since this pass
covers every item rather than only strict failures), and the resulting judge-augmented accuracy
(Appendix~\ref{app:regimetable} adds the judge-only figure). We avoid ``false negative''/``false
positive'': those terms presuppose the judge is ground truth, which it is not. On the 874 items
Gemini 2.5 Flash originally judged, phi4-mini agrees with it on 698 (\textbf{79.9\% raw agreement})---
neither suspiciously high (rubber-stamping) nor low enough to call either judge unreliable, but real,
bounded disagreement between two models neither of which we treat as ground truth. To resolve that
disagreement and check whether it tracks human judgment, we hand-adjudicate 238 items: the 174
REG/NUM/TMP items where Gemini and phi4-mini disagree, plus a stratified 64-item control sample where
they agree (Appendix~\ref{app:judge} gives the full sampling protocol). Adjudication is scored
against the judge verdicts as they stood when the sample was drawn, so every disagreement item
resolves to exactly one judge by construction; a later data-quality fix (Section~\ref{sec:models})
that corrected truncated reference text for a few items is reported independently rather than folded
in here. The adjudicator concurred with consensus on 89.1\% of control items; on the 174 disagreement
items, the human matched Gemini on 56.9\% and phi4-mini on 43.1\%, widest on numerical reasoning
(Appendix~\ref{app:judge} has the per-task breakdown). Neither judge is right often enough on these
disputed cases to be ground truth, and phi4-mini's lower share does not argue against using it as the
primary judge---that choice rests on full coverage, not on accuracy over Gemini.
Section~\ref{sec:limitations} returns to this as a limitation of the cross-model judge design.

\textbf{A pre-registered decision rule met an ambiguity it did not anticipate.} We had committed to
reporting DeepSeek-R1-Distill's strict-to-top reversal as a central finding only if the cross-model
judge supported it, and letting discriminative coverage (Section~\ref{sec:effsize}) carry more
weight otherwise. The rule presupposed one judge regime; there are two, and they disagree---the
reversal holds under judge-augmented scoring (Section~\ref{sec:regime}) but not judge-only, where
DeepSeek ranks 7th of twelve, tied with Qwen3-32B. Rather than picking the reading the rule
anticipated, we report both: the reversal is itself evidence for the paper's thesis that which of
two defensible ways of applying a judge changes the answer, and discriminative coverage is a
coequal finding, not a fallback.

\textbf{What survives the judge audit.} Attributing every strict failure to a task-conditioned
category (domain-knowledge, numerical, temporal, or context-grounding; Appendix~\ref{sec:errortax-sub})
shows the reclassification is uneven across models, not uniform. DeepSeek-R1-Distill's 101 strict
failures across all four tasks (99 on REG/NUM/TMP, the figure behind the 93.9\% reclassification rate
in Section~\ref{sec:regime}, plus 2 on CON, which is never judge-reviewed) collapse to 8 residual---the
largest correction of any model and the basis for its judge-augmented reversal---while Gemma 3~4B,
despite a comparable 72.1\% reclassification rate,
keeps a residual error rate 1.7--3$\times$ every other profiled model's. Appendix~\ref{sec:errortax-sub}
reports the full per-model breakdown
and representative failure examples.

\subsection{Statistical Significance and Performance Tiers}
\label{sec:significance}

Paired bootstrap significance testing (100,000 resamples, stable across 5 seeds) across all 66
model pairs finds 36 pairs significantly different at $p < 0.05$, of which 15--16 survive Bonferroni
correction ($\alpha = 0.05/66 \approx 0.00076$)---16 under 4 of 5 seeds, 15 under the fifth, because
one pair (DeepSeek-R1-Distill vs.\ Llama 4 Scout 17B) sits within Monte Carlo noise of the threshold
itself ($p \approx 0.0007$--$0.0008$); we report the range rather than a single seed's count for
that reason. Three descriptive performance clusters
organize the results under strict scoring; their boundaries are robust at the top and bottom of the
ranking but statistical, not categorical, in the middle.

\textbf{Tier 1 --- strong performers (81.5--89.7\%):} Gemini 2.5 Flash, Qwen3-32B, LLaMA-3.3-70B,
Llama 4 Scout 17B, and Kimi K2, all four of the latter statistically indistinguishable from one
another ($p$ = 0.06--0.79), forming a genuine plateau rather than a fine-grained ranking.
\textbf{The remaining seven models (70--79\%)} do not separate into a clean second and third tier:
Gemma 3 4B (78.8\%) is significantly below only the five Tier 1 models and is statistically
indistinguishable from every other model in the field ($p$ = 0.21--0.73), including
DeepSeek-R1-Distill, Mistral-7B, Gemini 2.5 Pro, and both GPT-OSS variants. Under strict scoring the
bottom seven models form one broad, largely undifferentiated cluster rather than the two-tier
structure a cursory read of the ranking would suggest.

Section~\ref{sec:discussion} returns to two Tier 1 pairs where a large parameter difference produces
no measurable accuracy difference.

\textbf{Robustness to document-level clustering.} 406 items are not 406 independent draws: they trace
to 34 source documents (Section~\ref{sec:models}, Section~\ref{app:provenance}), and several
contradiction-detection items link two documents at once. We re-ran \emph{all 66} pairwise
comparisons under a document-clustered bootstrap---resampling the 27 connected components formed by
unioning documents that co-occur in any contradiction item, rather than resampling items
directly---not only the comparisons this paper singles out as findings, since checking only
hand-picked pairs would risk an unrepresentatively reassuring picture. The clustered test finds
\textbf{29 of 66} pairs significant at $p<0.05$ (item-level: 36) and \textbf{11} surviving Bonferroni
correction (item-level: 15--16, itself boundary-sensitive to Monte Carlo noise across independent
10,000-resample runs). \textbf{Nine pairs flip significance direction} under clustering---all among
comparisons this paper does not otherwise discuss. Every comparison this paper specifically reports
as a finding, including the human-reference comparison in Section~\ref{sec:human}, agrees in
direction between the two tests: the correction changes the overall significance count meaningfully
(36$\to$29) without overturning any conclusion this paper draws. Section~\ref{app:clustering}
reports the full 66-pair comparison and the clustering construction.

\begin{table}[h]
\centering\small
\begin{tabular}{lrrr}
\toprule
Model & Overturn \% & Rejection \% & 406-item judge-augmented acc. \\
\midrule
Gemini 2.5 Flash & 80.0 & 9.1 & 96.5 \\
Qwen3-32B & 90.6 & 6.9 & 97.3 \\
LLaMA-3.3-70B & 92.1 & 13.5 & 98.0 \\
Llama 4 Scout 17B & 86.6 & 4.7 & 97.5 \\
Kimi K2 & 94.3 & 6.6 & 97.8 \\
Gemma 3 4B & 84.9 & 10.3 & 94.1 \\
LLaMA-3-8B & 87.1 & 4.2 & 96.3 \\
GPT-OSS 120B & 92.2 & 5.9 & 97.5 \\
GPT-OSS 20B & 92.3 & 7.1 & 97.5 \\
Gemini 2.5 Pro & 89.2 & 2.8 & 96.5 \\
Mistral-7B & 93.0 & 6.6 & 95.6 \\
DeepSeek-R1-Distill & 93.9 & 9.4 & 98.0 \\
\bottomrule
\end{tabular}
\caption{Cross-model judge (phi4-mini) reliability by model, full REG/NUM/TMP coverage (344 items
each). Overturn \% is the strict-to-judge overturn rate (strict-wrong items the judge marks
correct); Rejection \% is the judge's rejection rate of strict-correct predictions. Neither is a
false-negative/false-positive rate against ground truth---phi4-mini is not treated as one (see
Section~\ref{sec:judge}). Final column is 406-item accuracy including CON, under the
judge-augmented (strict-OR-judge) regime; Appendix~\ref{app:regimetable} reports the same models'
406-item accuracy under the judge-only regime, where the judge's verdict is final in both
directions.}
\label{tab:judgereliability}
\end{table}


\section{Discussion}
\label{sec:discussion}

\subsection{What a Single Accuracy Number Hides}

Strict and judge-only rank the same twelve models on the same 406 items, and they disagree about the
ordering in both directions, not just one. Gemini 2.5 Pro moves from 10th under strict scoring to
1st under the judge's own verdict (+18.7 points, 9 ranks) --- the single largest mover in the study.
LLaMA-3.3-70B moves the other way, from 3rd to 10th (a 4.9-point gain that costs it 7 ranks, because
every other model gained more). Gemini 2.5 Flash's raw score is identical under both regimes (89.7\%
either way) yet it still falls from 1st to 9th, because the rest of the field moved past it. That
disagreement is not noise to be averaged
away; it is the measurement telling us that two distinct quantities have been collapsed into one.
Strict accuracy measures what a format-sensitive pipeline can extract. The judge's verdict measures
a second model's judgment of substantive correctness, which is its own imperfect measurement, not a
ground truth the strict number falls short of.

The practical consequence is that a leaderboard cannot be read without knowing which question it
answers. If the intended use is an automated compliance pipeline that parses model output
mechanically, the strict number is the relevant one. If the question is which model a second,
independent-family model judges most likely to be substantively correct, the judge-only number is
the relevant one, and for several models in this study the two rankings point in different
directions. Benchmarks that report one figure force the reader to guess which of these they are
looking at.

We would not argue for replacing string matching with judge-based scoring. Judges have their own
failure modes, and ours is no exception: the Gemini pilot's 28-item manual audit found every error
running the same direction (Appendix~\ref{app:judge})---over-crediting a model that showed sound
intermediate work but reached a wrong final answer. Whether that bias systematically favours verbose
or reasoning-heavy models specifically is a claim we tested rather than asserted (Section~\ref{sec:regime}),
and it does not survive adjustment for task and model composition. What we can say is only that the
two regimes disagree, sometimes sharply, in both directions, which is the argument for reporting
both and treating the interval between them as the honest statement of what is known.

\subsection{Scale Does Not Explain These Results}

Of the 66 pairwise comparisons, 15--16 survive Bonferroni correction, seed-sweep-verified
(Section~\ref{sec:significance}).
The surviving differences separate the top of the table from the bottom; the middle is largely
undifferentiated, and two comparisons make the point sharply.

Llama 4 Scout 17B and LLaMA-3.3-70B are statistically indistinguishable (83.3\% vs 83.7\%,
$p = 0.786$) across a four-fold parameter difference. GPT-OSS 120B and GPT-OSS 20B are likewise
indistinguishable (77.1\% vs 76.8\%, $p = 0.912$) across a six-fold difference within a single model
family. The second case admits a stronger statement than equal accuracy: the two GPT-OSS models have
the highest error overlap of any pair in the study (Jaccard 0.427 against a median of 0.274 across
all 66 pairs). They do not merely score alike---they fail on substantially the same items. Whatever
limits them on this benchmark is a property the additional parameters did not address.

\subsection{What This Means for Benchmark Construction}

Two findings here were not things we set out to measure, and both concern the benchmark rather than
the models.

The first is that over a third of our items do not discriminate at all. 148 of 406 (36.5\%) are
answered identically by all twelve models. We built the benchmark for coverage of a regulatory domain, which is a reasonable
objective, but coverage and discrimination are different targets and expert annotation effort spent
on one does not buy the other. A benchmark reporting only nominal size overstates the evidence
behind any leaderboard gap it produces.

The second concerns our difficulty labels. We assigned difficulty by counting reasoning steps, and
LLaMA-3.3-70B performs \emph{better} on our hard items than our easy ones. The likeliest reading is
that reasoning-step count captures one axis of difficulty while missing another---hard items here
tend to state their amendment-chain cues explicitly, whereas medium items turn on a single subtle
qualifier. We report the label scheme's failure rather than repairing it after the fact, because the
inversion is informative about what makes regulatory text hard.

\section{Conclusion}
\label{sec:conclusion}

Benchmark accuracy is reported as a property of a model. It is a property of a model and a scoring
rule together, and on IndiaFinBench that joint dependence is large enough to change which model looks
best. Scoring twelve LLMs under strict string matching and under a full-coverage cross-model judge's
own verdict, over the same predictions, produces two rankings with no positive correspondence
(Spearman $\rho = -0.273$): some models score higher under the judge and rank higher, some score
higher under the judge and rank \emph{lower} because the field moved past them, and one model's raw
score does not move at all yet its rank falls by eight places. Applying the judge asymmetrically---to
rescue strict failures but never to challenge strict successes---produces a third ranking again,
under which the model placed last under strict scoring ties for first, because 93.9\% of its strict
errors are reclassified as correct. Neither regime is ground truth, and neither is universally
preferable; they answer different questions, and
reporting one alone leaves the reader unable to tell which.

Underneath that, over a third of the benchmark carries no discriminative signal: 148 of 406 items
(36.5\%) are answered identically by all twelve models, while a predefined higher-discrimination
band captures 134 items that separate systems more sharply. Nominal size and discriminative
coverage are different quantities, and the difference bounds how much weight any reported gap can
carry.

IndiaFinBench is released in full---406 expert-annotated items over 34 of 192 collected SEBI and RBI documents,
the evaluation harness, per-item judge verdicts, and all model predictions---so that all three
regimes can be reproduced and any of their claims contested. The context-only setup isolates
inference over supplied regulatory passages from recall of memorised training text, although it
cannot rule out training-data exposure to the underlying documents; the Indian regulatory setting
makes this isolation meaningful because the source text is jurisdiction-specific and less likely to
dominate pretraining corpora than US or European financial text. Whether this degree of
scoring-regime sensitivity generalizes beyond Indian financial regulation is an empirical question
for future evaluations; our results motivate reporting more than one regime on other benchmarks.


\section*{Limitations}
\label{sec:limitations}

Several limitations should be noted. The automated four-stage scoring pipeline penalizes
semantically correct predictions that differ in number format or surface phrasing; an early
single-judge pilot (Gemini 2.5 Flash, 874 items, Appendix~\ref{app:judge}) reclassified 78.4\% of
flagged items as correct. None of strict,
judge-only, or judge-augmented scoring is ground truth: strict and the judge's own verdict have
opposing error profiles, and a second, cross-model
judge (Section~\ref{sec:judge}) only partially addresses this, since sharing no model family does not
rule out both judges sharing a bias human adjudicators would not. Human adjudication
(Section~\ref{sec:judge}) confirms this concern is not hypothetical: on the 174 disputed items,
phi4-mini---this paper's primary judge, chosen for full coverage rather than higher accuracy on
disputed cases---is the less accurate of the two judges specifically where they disagree.
DeepSeek-R1-Distill has 14 items in this disputed sample (of its 99 strict REG/NUM/TMP errors);
phi4-mini reclassified 10 of these 14 as correct, and human review disagreed with 7 of those 10. This
is a small, disagreement-selected subsample---not sized to estimate what fraction of DeepSeek's
93.9\% reclassification rate would survive full human review---but it is a genuine limitation of the
cross-model judge design worth stating plainly rather than only reporting the more favourable
control-agreement rate (89.1\%). Because this adjudication was performed by the author, two
independent, non-author adjudicators separately completed a 62-item stratified subsample of the same
disagreement set (Appendix~\ref{app:independent-adjudication}); their pairwise agreement with each
other and with the author's own verdicts falls in the substantial-agreement band ($\kappa=0.65$--$0.74$),
and both independently reproduce the same near-even Gemini/phi4-mini split the author found alone,
which is evidence against the specific concern that the author's own adjudication is idiosyncratic,
though it does not extend to a larger or differently-selected sample. The benchmark also does not
cover Hindi--English code-switched regulatory text that appears in some official documents.

The human reference study involves a single non-specialist evaluator on 60 medium-and-hard items; it
anchors the harder half of the benchmark but supports no population-level claims about human
performance, and a domain specialist would likely score higher. The IAA study covers 180 of 406
items (44.3\%) across three rounds, 86.1\% overall agreement with CON at $\kappa$ = 0.712; agreement on
open-ended tasks reflects the same surface-form scoring sensitivity documented in Appendix~\ref{app:judge}.
Few-shot evaluation was conducted on only four top-performing models (Appendix~\ref{app:fewshot}).

One checkpoint identity rests on inference rather than a surviving primary record: Gemma's local
Ollama tag was re-confirmed as \texttt{gemma3:4b} via the API's own model-metadata endpoint
(4,299,915,632 parameters, family \texttt{gemma3}) after an initial ambiguity in the evaluation
script, and this matches the parameter count already stated in our own methods text---which is
consistent evidence, not fully independent confirmation, since the methods text and the re-run
identity could in principle have been aligned to each other rather than to an external record. We
consider this resolved with reasonable confidence but flag the residual uncertainty rather than
overstating it as settled fact.

A structural limitation shared across all evaluations is context-injection: models receive the
relevant passage directly rather than retrieving it from the full document corpus, testing reading
comprehension over a provided excerpt rather than the retrieval or full-document understanding
real-world compliance workflows also require. The benchmark is a static snapshot of SEBI and RBI
regulation as of early 2026 and will need periodic refresh as the regulatory landscape evolves.

Primary annotation was conducted by a single domain-expert annotator, with the independent human
inter-annotator study above providing external validation; Section~\nameref{sec:humansubjects}
below states who the annotators and evaluator were and what oversight, if any, applied. All
questions and gold-standard answers are in English; the official Hindi-language versions of the same
circulars are not included, limiting applicability to multilingual regulatory analysis. Coverage is
also limited to two regulators---extending to the Insurance Regulatory and Development Authority of
India (IRDAI), the Pension Fund Regulatory and Development Authority (PFRDA), and commodity-segment
regulation would substantially broaden the benchmark's coverage of the Indian financial regulatory
ecosystem.

Finally, IndiaFinBench evaluates short extractive responses rather than longer-form generation; it
does not test whether models can draft regulatory summaries, identify compliance gaps, or produce
legally coherent reasoning chains---capabilities that are arguably more consequential for
legal-technology applications and a natural next step for the benchmark.

We also note a construction limitation surfaced by our own analysis rather than external review: the
difficulty label assigned at annotation time (Section~\ref{sec:models}) predicts empirical
difficulty for only 3 of the 12 evaluated models (Section~\ref{sec:difficulty}), and roughly half of
all items do not discriminate between the systems evaluated here (Section~\ref{sec:effsize}). Both
findings bear on how confidently any single-digit accuracy gap on this benchmark, ours or a future
system's, should be read.

\section*{Ethics Statement}
\label{sec:humansubjects}

\textbf{Source data.} IndiaFinBench is built from primary source documents published by the
Securities and Exchange Board of India (sebi.gov.in) and the Reserve Bank of India (rbi.org.in) for
public access. These are works of an Indian government body; we make no claim that they are in the
public domain or free of copyright, and we do not represent the CC~BY~4.0 license below as covering
the source documents themselves. What is released under CC~BY~4.0 is the authors' own contribution:
the questions, reference answers, difficulty and task-type annotations, evaluation harness, and
model predictions. Every item's provenance (source document and, where applicable, URL) is included
in the release so that claims can be traced back to the original regulatory text.

\textbf{Human participants.} Four human contributions appear in this paper: the second annotator in
the inter-annotator agreement study (Section~\ref{sec:iaa}), the non-specialist evaluator
(Section~\ref{sec:human}), and two independent adjudicators (Appendix~\ref{app:independent-adjudication}).
All were volunteers from the author's personal and academic network (fellow students and family
members), recruited directly rather than through any institutional subject pool, informed of the
study's purpose before participating, and uncompensated. No demographic, identifying, or personal
information about any participant is collected, stored, or released; the public dataset contains
only their task responses (answers to benchmark items, or adjudication verdicts), not any
information about who they are, and the two independent adjudicators are referred to only as
Adjudicator A and Adjudicator B throughout. \textbf{Institutional review status:} this activity has not
been reviewed by an institutional ethics board, and we make no determination here about whether such
review would be required under any particular institution's policy---that determination is left to
readers applying their own institution's standards, and to the author's institution before any
subsequent submission of this or related work. We report this status plainly rather than asserting
an exemption we have not confirmed.

\textbf{Content and use.} The benchmark is designed to evaluate model performance on regulatory
reasoning tasks and does not contain toxic, harmful, or privacy-violating content. We are not aware
of a use of this benchmark, or of the models evaluated against it, that would constitute a
foreseeable harm beyond the ordinary risks of applying any LLM evaluation to a regulated domain: a
practitioner treating any model's output here as a substitute for qualified legal or compliance
advice would be misusing the benchmark's findings, which concern relative model performance on
constructed QA items, not the models' fitness for real regulatory decision-making.

\subsubsection*{Broader Impact Statement}

This work measures how scoring methodology affects reported LLM performance and releases a new
evaluation resource; it does not train or release a new model, and we see no material risk of
negative societal impact beyond the general risks of LLM benchmarking research. To the extent this
paper has a broader implication, it argues for more transparent, multi-regime benchmark reporting,
which we expect to reduce rather than increase the risk of over-trusting a single leaderboard number
in deployment decisions.

\label{sec:mainend}
\bibliography{indiafinbench}

@article{ho2025reassessing,
  title   = {Reassessing Extractive {QA} Datasets at Scale: {LLM}-as-a-Judge and In-Depth Analyses},
  author  = {Ho, Xanh and Huang, Jiahao and Boudin, Florian and Aizawa, Akiko},
  journal = {arXiv preprint arXiv:2504.11972},
  year    = {2025}
}

@article{norman2026reliability,
  title   = {Reliability without Validity: A Systematic, Large-Scale Evaluation of {LLM}-as-a-Judge Models Across Agreement, Consistency, and Bias},
  author  = {Norman, Justin D. and Rivera, Michael U. and Hughes, D. Alex},
  journal = {arXiv preprint arXiv:2606.19544},
  year    = {2026}
}

@inproceedings{chen2021finqa,
  title     = {{F}in{QA}: A Dataset of Numerical Reasoning over Financial Data},
  author    = {Chen, Zhiyu and Chen, Wenhu and Smiley, Charese and Shah, Sameena and Borova, Iana and Langdon, Dylan and Moussa, Reema and Beane, Matt and Huang, Ting-Hao and Routledge, Bryan and Wang, William Yang},
  booktitle = {Proceedings of the 2021 Conference on Empirical Methods in Natural Language Processing (EMNLP)},
  year      = {2021}
}

@inproceedings{zheng2022convfinqa,
  title     = {{C}onv{F}in{QA}: Exploring the Chain of Numerical Reasoning in Conversational Finance Question Answering},
  author    = {Chen, Zhiyu and Li, Shiyang and Smiley, Charese and Ma, Zhiqiang and Shah, Sameena and Wang, William Yang},
  booktitle = {Proceedings of the 2022 Conference on Empirical Methods in Natural Language Processing (EMNLP)},
  year      = {2022}
}

@inproceedings{loukas2022finer,
  title     = {{F}i{NER}-139: A Dataset for Fine-Grained Named Entity Recognition in Financial Text},
  author    = {Loukas, Lefteris and Fergadiotis, Manos and Chalkidis, Ilias and Spyropoulou, Eirini and Malakasiotis, Prodromos and Androutsopoulos, Ion and Paliouras, Georgios},
  booktitle = {Proceedings of the 2022 Conference on Empirical Methods in Natural Language Processing (EMNLP)},
  year      = {2022}
}

@inproceedings{shah2022flue,
  title     = {{FLUE}: Financial Language Understanding Evaluation},
  author    = {Shah, Raj Sanjay and Chawla, Kunal and Eyer, Dheeraj and Solorio, Thamar and Yin, Wei},
  booktitle = {Findings of the Association for Computational Linguistics: EMNLP 2022},
  year      = {2022}
}

@inproceedings{xie2023pixiu,
  title     = {{PIXIU}: A Large Language Model, Instruction Data and Evaluation Benchmark for Finance},
  author    = {Xie, Qianqian and Han, Weiguang and Zhang, Xiao and Lai, Yanzhao and Peng, Min and Lopez-Lira, Alejandro and Huang, Jimin},
  booktitle = {Advances in Neural Information Processing Systems 36 (NeurIPS 2023 Datasets and Benchmarks Track)},
  year      = {2023}
}

@inproceedings{xie2024finben,
  title     = {{F}in{B}en: A Holistic Financial Benchmark for Large Language Models},
  author    = {Xie, Qianqian and Han, Weiguang and Chen, Zhengyu and Xiang, Ruoyu and Zhang, Xiao and He, Yueru and Xiao, Mengxi and Li, Dong and Dai, Yongfu and Feng, Duanyu and Xu, Yijing and Kang, Haoqiang and Kuang, Ziyan and Yuan, Chenhan and Yang, Kailai and Luo, Zheheng and Zhang, Tianlin and Liu, Zhiwei and Xiong, Guojun and Deng, Zhiyang and Jiang, Yuechen and Yao, Zhiyuan and Li, Haohang and Yu, Yangyang and Hu, Gang and Huang, Jimin and Liu, Xiao-Yang and Lopez-Lira, Alejandro and Wang, Benyou and Lai, Yanzhao and Wang, Hao and Peng, Min and Ananiadou, Sophia and Huang, Jian-Yun},
  booktitle = {Advances in Neural Information Processing Systems 37 (NeurIPS 2024 Datasets and Benchmarks Track)},
  year      = {2024}
}

@article{islam2023financebench,
  title   = {{F}inance{B}ench: A New Benchmark for Financial Question Answering},
  author  = {Islam, Pranab and Kannappan, Anand and Kiela, Douwe and Qian, Rebecca and Scherrer, Nino and Vidgen, Bertie},
  journal = {arXiv preprint arXiv:2311.11944},
  year    = {2023}
}

@inproceedings{hendrycks2021cuad,
  title     = {{CUAD}: An Expert-Annotated {NLP} Dataset for Legal Contract Review},
  author    = {Hendrycks, Dan and Burns, Collin and Chen, Anya and Ball, Spencer},
  booktitle = {Advances in Neural Information Processing Systems 34 (NeurIPS 2021 Datasets and Benchmarks Track)},
  year      = {2021}
}

@inproceedings{chalkidis2022lexglue,
  title     = {Lex{GLUE}: A Benchmark Dataset for Legal Language Understanding in {E}nglish},
  author    = {Chalkidis, Ilias and Jana, Abhik and Hartung, Dirk and Bommarito, Michael and Androutsopoulos, Ion and Katz, Daniel Martin and Aletras, Nikolaos},
  booktitle = {Proceedings of the 60th Annual Meeting of the Association for Computational Linguistics (Volume 1: Long Papers)},
  year      = {2022}
}

@inproceedings{malik2021ildc,
  title     = {{ILDC} for {CJPE}: Indian Legal Documents Corpus for Court Judgment Prediction and Explanation},
  author    = {Malik, Vijit and Sanjay, Rishabh and Nigam, Shubham Kumar and Ghosh, Kripabandhu and Guha, Shouvik Kumar and Bhattacharya, Arnab and Modi, Ashutosh},
  booktitle = {Proceedings of the 59th Annual Meeting of the Association for Computational Linguistics and the 11th International Joint Conference on Natural Language Processing (Volume 1: Long Papers)},
  year      = {2021}
}

@inproceedings{liang2022helm,
  title     = {Holistic Evaluation of Language Models},
  author    = {Liang, Percy and Bommasani, Rishi and Lee, Tony and Tsipras, Dimitris and Soylu, Dilara and Yasunaga, Michihiro and Zhang, Yian and Narayanan, Deepak and Wu, Yuhuai and Kumar, Ananya and others},
  booktitle = {Advances in Neural Information Processing Systems 35 (NeurIPS 2022)},
  year      = {2022}
}

@inproceedings{rajpurkar2016squad,
  title     = {{SQ}u{AD}: 100,000+ Questions for Machine Comprehension of Text},
  author    = {Rajpurkar, Pranav and Zhang, Jian and Lopyrev, Konstantin and Liang, Percy},
  booktitle = {Proceedings of the 2016 Conference on Empirical Methods in Natural Language Processing (EMNLP)},
  year      = {2016}
}

@inproceedings{dua2019drop,
  title     = {{DROP}: A Reading Comprehension Benchmark Requiring Discrete Reasoning Over Paragraphs},
  author    = {Dua, Dheeru and Wang, Yizhong and Dasigi, Pradeep and Stanovsky, Gabriel and Singh, Sameer and Gardner, Matt},
  booktitle = {Proceedings of the 2019 Conference of the North American Chapter of the Association for Computational Linguistics: Human Language Technologies (Volume 1: Long and Short Papers)},
  year      = {2019}
}

@article{landiskoch1977,
  title   = {The Measurement of Observer Agreement for Categorical Data},
  author  = {Landis, J. Richard and Koch, Gary G.},
  journal = {Biometrics},
  volume  = {33},
  number  = {1},
  pages   = {159--174},
  year    = {1977}
}

@article{wilson1927,
  title   = {Probable Inference, the Law of Succession, and Statistical Inference},
  author  = {Wilson, Edwin B.},
  journal = {Journal of the American Statistical Association},
  volume  = {22},
  number  = {158},
  pages   = {209--212},
  year    = {1927}
}

@inproceedings{zheng2023judging,
  title     = {Judging {LLM}-as-a-Judge with {MT}-Bench and Chatbot Arena},
  author    = {Zheng, Lianmin and Chiang, Wei-Lin and Sheng, Ying and Zhuang, Siyuan and Wu, Zhanghao and Zhuang, Yonghao and Lin, Zi and Li, Zhuohan and Li, Dacheng and Xing, Eric P. and Zhang, Hao and Gonzalez, Joseph E. and Stoica, Ion},
  booktitle = {Advances in Neural Information Processing Systems 36 (NeurIPS 2023 Datasets and Benchmarks Track)},
  year      = {2023}
}

@inproceedings{wang2024fairevaluators,
  title     = {Large Language Models are not Fair Evaluators},
  author    = {Wang, Peiyi and Li, Lei and Chen, Liang and Cai, Zefan and Zhu, Dawei and Lin, Binghuai and Cao, Yunbo and Kong, Lingpeng and Liu, Qi and Liu, Tianyu and Sui, Zhifang},
  booktitle = {Proceedings of the 62nd Annual Meeting of the Association for Computational Linguistics (Volume 1: Long Papers)},
  year      = {2024}
}

@article{wataoka2024selfpreference,
  title   = {Self-Preference Bias in {LLM}-as-a-Judge},
  author  = {Wataoka, Koki and Takahashi, Tsubasa and Ri, Ryokan},
  journal = {arXiv preprint arXiv:2410.21819},
  year    = {2024}
}

@article{gu2024survey,
  title   = {A Survey on {LLM}-as-a-Judge},
  author  = {Gu, Jiawei and Jiang, Xuhui and Shi, Zhichao and Tan, Hexiang and Zhai, Xuehao and Xu, Chengjin and Li, Wei and Shen, Yinghan and Ma, Shengjie and Liu, Honghao and Wang, Saizhuo and Zhang, Kun and Wang, Yuanzhuo and Gao, Wen and Ni, Lionel and Guo, Jian},
  journal = {arXiv preprint arXiv:2411.15594},
  year    = {2024}
}
\bibliographystyle{tmlr}

\appendix

\section{Document Provenance and Clustered Significance Testing}
\label{app:provenance}
\label{app:clustering}

\subsection{Full Scoring-Regime Comparison}
\label{app:regimetable}

Table~\ref{tab:regime} tabulates exact accuracy and rank under all three scoring regimes (strict,
judge-only, judge-augmented) for all twelve models; Section~\ref{sec:regime} and
Figure~\ref{fig:regime} discuss the comparisons that matter.

\begin{table}[h]
\centering
\caption{Accuracy and rank under all three scoring regimes. Strict is the automated string-matching
pipeline; judge-only treats the cross-model judge's verdict as final in both directions;
judge-augmented is the asymmetric composite strict-OR-judge (a sensitivity regime, not a primary
one). Ranks are competition ranks, so tied scores share the better rank. Generated from the released
per-item judge verdicts by \texttt{scripts/regime\_table.py}.}
\label{tab:regime}
\begin{tabular}{lrrrrrr}
\toprule
Model & Strict & Rank & Judge-only & Rank & Judge-augmented & Rank \\
\midrule
Gemini 2.5 Flash & 89.66 & 1 & 89.66 & 9 & 96.55 & 5 \\
Qwen3-32B & 85.47 & 2 & 92.36 & 7 & 97.29 & 4 \\
LLaMA-3.3-70B & 83.74 & 3 & 88.67 & 10 & 98.03 & 1 \\
Llama 4 Scout 17B & 83.25 & 4 & 94.33 & 2 & 97.54 & 3 \\
Kimi K2 & 81.53 & 5 & 93.35 & 5 & 97.78 & 2 \\
Gemma 3 4B & 78.82 & 6 & 87.19 & 11 & 94.09 & 8 \\
LLaMA-3-8B & 78.08 & 7 & 93.6 & 4 & 96.31 & 6 \\
GPT-OSS 120B & 77.09 & 8 & 93.84 & 3 & 97.54 & 3 \\
GPT-OSS 20B & 76.85 & 9 & 93.1 & 6 & 97.54 & 3 \\
Gemini 2.5 Pro & 76.11 & 10 & 94.83 & 1 & 96.55 & 5 \\
Mistral-7B & 75.86 & 11 & 91.38 & 8 & 95.57 & 7 \\
\textbf{DeepSeek-R1-Distill} & \textbf{75.12} & \textbf{12} & \textbf{92.36} & \textbf{7} & \textbf{98.03} & \textbf{1} \\
\bottomrule
\end{tabular}

\end{table}

\subsection{Document Provenance}

The 192 documents listed in Appendix~\ref{app:sourcedocs} are the full collected corpus. The 406
annotated items trace to \textbf{34} of them, spanning both regulators; Table~\ref{tab:provenance}
summarises the distribution.

\begin{table}[h]
\centering\small
\begin{tabular}{lr}
\toprule
Documents collected (full corpus) & 192 \\
Documents represented in the 406 items & 34 \\
Largest single document (item count) & 63 \\
Median items per document & 6 \\
\bottomrule
\end{tabular}
\caption{Document provenance summary. Full item-level provenance
(\texttt{evaluation/provenance\_audit.csv}) is released with the benchmark.}
\label{tab:provenance}
\end{table}

\subsection{Document-Clustered Bootstrap}

Items sharing a source document are not independent draws. We construct resampling clusters as the
connected components of a graph over the 34 documents, with an edge added between
\texttt{document\_a} and \texttt{document\_b} for every contradiction-detection item---since a CON
item depends on two documents jointly, keeping them in separate clusters would let dependence leak
across the resampling unit boundary. This produced \textbf{27} connected components (sizes: 68, 48,
46, 41, 41, 35, 20, 13, 12, 11, 11, 10, 9, 7, 6, 4, 4, 4, 3, 3, 2, 2, 2, 1, 1, 1, 1); four are singleton
documents that appear in no contradiction pair.

We re-ran paired-bootstrap significance testing (100,000 resamples, stable across 5 seeds --
scripts/bootstrap\_seed\_sweep.py) at the document-cluster level for
\textbf{all 66 pairwise comparisons}, plus the human-reference comparison separately: resampling
clusters with replacement, and including every item in each drawn cluster's data, rather than
resampling items directly. Table~\ref{tab:clustered} reports the seven comparisons discussed in the
main text; the full 66-pair table is released with the evaluation code
(\texttt{evaluation/clustered\_bootstrap\_full66.json}).

\begin{table}[h]
\centering\small
\begin{tabular}{lrr}
\toprule
Comparison & Item-level $p$ & Cluster-level $p$ \\
\midrule
Gemini 2.5 Flash vs.\ Qwen3-32B & 0.052 & 0.060 \\
Gemini 2.5 Flash vs.\ Gemma 3 4B & $<$0.001 & $<$0.001 \\
Llama 4 Scout 17B vs.\ LLaMA-3.3-70B & 0.788 & 0.825 \\
GPT-OSS 120B vs.\ GPT-OSS 20B & 0.904 & 0.936 \\
Gemma 3 4B vs.\ DeepSeek-R1-Distill & 0.186 & 0.424 \\
Gemma 3 4B vs.\ Mistral-7B & 0.254 & 0.431 \\
Human vs.\ Gemini 2.5 Flash (60-item subset, 17 clusters) & 0.427 & 0.558 \\
\bottomrule
\end{tabular}
\caption{Item-level vs.\ document-clustered bootstrap $p$-values for the seven comparisons discussed
in the main text. All seven agree in significance direction. Across the full 66-pair table, 29 of 66
are significant under clustering versus 36 item-level, and 9 pairs (not shown here, none discussed
in the main text) flip significance direction --- see the released full comparison for the complete
picture rather than only these seven.}
\label{tab:clustered}
\end{table}

Cluster assignment was checked for correctness before use: every item maps to exactly one component,
every document appears in exactly one component, and the procedure is deterministic under a fixed
random seed (verified by re-running with an identical seed and confirming bitwise-identical output).
We do not treat "the clustered interval is wider" as a correctness check---no such guarantee exists
in general--- and instead verified the construction directly.

\section{Error Analysis}
\label{sec:errortax-sub}

\subsection{Task-Aligned Error Attribution}

We attribute model failures to four categories via a deterministic mapping from each item's task
type and difficulty (released with the evaluation code): \textbf{Domain Knowledge Failure (DKF)}:
unfamiliarity with Indian regulatory concepts, terminology, or thresholds. \textbf{Numerical
Reasoning Failure (NRF)}: arithmetic errors---wrong calculation, unit conversion, or formula
application despite the formula appearing in context. \textbf{Temporal Reasoning Failure (TRF)}:
mis-ordering regulatory events, misidentifying the circular in force at a given time, or failing to
track supersession. \textbf{Context Grounding Failure (CGF)}: using external knowledge instead of
the passage, or failing to extract a span clearly present in context. Contradiction-detection errors
on medium and hard items map to TRF, since the task probes supersession tracking; easy CON items and
hard REG items both map to CGF. \textbf{These labels are task-conditioned categories, not
diagnoses of a model's internal failure mechanism}: because the mapping is deterministic in task
type and difficulty, an error's label carries no information beyond which task and difficulty tier
it occurred in. A pattern stated purely in terms of these labels (e.g.\ ``TRF is common'') is a
restatement of which task the errors fall in, not an independent finding about that task; the
findings below that carry independent information are the ones stated in terms of per-model
\emph{reclassification rates}, not error-type composition.

Table~\ref{tab:errortax} shows error distributions for five key models under both scoring regimes: strict-pipeline errors, and residual errors that survive the cross-model judge audit (phi4-mini, Section~\ref{sec:judge}, Appendix~\ref{app:judge}). The distinction matters because strict errors mix true failures with format mismatches, whereas residual errors are judge-retained.

\begin{table*}[t]
\centering\small
\begin{tabular}{llrrrrr}
\toprule
\textbf{Model} & \textbf{Errors} & \textbf{DKF} & \textbf{NRF} & \textbf{TRF} & \textbf{CGF} & \textbf{Total} \\
\midrule
\multirow{2}{*}{Gemini 2.5 Flash} & strict & 12 (29\%) & 14 (33\%) & 13 (31\%) & 3 (7\%) & 42 \\
 & residual & 4 (29\%) & 2 (14\%) & 7 (50\%) & 1 (7\%) & 14 \\
\multirow{2}{*}{Qwen3-32B} & strict & 25 (42\%) & 21 (36\%) & 10 (17\%) & 3 (5\%) & 59 \\
 & residual & 1 (9\%) & 3 (27\%) & 6 (55\%) & 1 (9\%) & 11 \\
\multirow{2}{*}{LLaMA-3.3-70B} & strict & 25 (38\%) & 23 (35\%) & 17 (26\%) & 1 (2\%) & 66 \\
 & residual & 0 (0\%) & 2 (25\%) & 5 (62\%) & 1 (12\%) & 8 \\
\multirow{2}{*}{DeepSeek-R1-Distill-Llama-70B} & strict & 51 (50\%) & 28 (28\%) & 19 (19\%) & 3 (3\%) & 101 \\
 & residual & 2 (25\%) & 1 (12\%) & 5 (62\%) & 0 (0\%) & 8 \\
\multirow{2}{*}{Gemma 3 4B} & strict & 25 (29\%) & 27 (31\%) & 31 (36\%) & 3 (3\%) & 86 \\
 & residual & 1 (4\%) & 6 (25\%) & 15 (62\%) & 2 (8\%) & 24 \\
\bottomrule
\end{tabular}

\caption{Error distribution by type for five key models under strict scoring and after cross-model judge audit (residual). DKF: Domain Knowledge Failure; NRF: Numerical Reasoning Failure; TRF: Temporal Reasoning Failure; CGF: Context Grounding Failure. Models selected as top performer (Gemini 2.5 Flash), bottom performer (Gemma 3 4B), and three distinctive profiles. Counts derive from the released rule-based mapping and per-item judge verdicts; CON errors are not judge-reviewed, so they persist into the residual column unconditionally.}
\label{tab:errortax}
\end{table*}

Three patterns carry independent information, stated as per-model reclassification rates rather
than error-type composition. First, DeepSeek-R1-Distill-Llama-70B shows the largest correction: 101
strict failures (across REG, NUM, TMP, and CON) collapse to 8 residual---6 of which are the
REG/NUM/TMP residual failures behind the 93.9\% reclassification figure in
Section~\ref{sec:regime}, the remaining 2 are CON errors, which are never judge-reviewed---the
largest reclassification of any profiled model and the basis for its rank reversal. Second, Gemma
3 4B's errors also fall substantially under judging (86 $\rightarrow$ 24, a 72.1\% reclassification
rate), so its strict-error count no longer supports a claim that its errors are format-invariant;
what does survive is its residual \emph{rate}: at 24/406 = 5.9\% of all items, it remains
1.7--3$\times$ every other profiled model's residual rate (2.0--3.4\%). Third, Gemini
2.5 Flash---the top strict performer---has the \emph{lowest} reclassification rate of the five
(66.7\%): proportionally more of its strict failures are judge-retained as genuine, consistent with
its output already being well-formatted under strict scoring, leaving fewer format artifacts
available to reclassify in the first place.

A task-level pattern is also visible in the residual composition, with the caveat above that it
restates task and difficulty rather than adding independent information: items whose deterministic
label is TRF (chiefly temporal-reasoning items, plus medium/hard contradiction-detection items) make
up 50\% (Gemini 2.5 Flash: 7/14) to 62\% (LLaMA-3.3-70B: 5/8, DeepSeek-R1-Distill-Llama-70B: 5/8,
Gemma 3 4B: 15/24) of every profiled model's residual, judge-retained errors, while items labelled
NRF or DKF are cleared at a substantially higher rate under judging. Read as a task-level fact rather
than a mechanistic one, this says supersession-tracking items resist the judge's format-tolerance
more than numerical or domain-knowledge items do, for every model profiled here --- not that models
share a specific ``temporal reasoning'' failure mode. Gemma 3 4B's residual concentration noted above
is, by the same deterministic mapping, concentrated in items labelled NRF (6 of its 24 residual
errors, the largest absolute NRF count of the five). Context Grounding Failure is rare under both
regimes for every model.

Representative failure examples for each error type are given in Appendix~\ref{app:examples}.

\section{Source Document Categories}
\label{app:sourcedocs}

\textbf{SEBI Documents:} SEBI (Issue of Capital and Disclosure Requirements) Regulations 2018, SEBI (Listing Obligations and Disclosure Requirements) Regulations 2015, SEBI (Substantial Acquisition of Shares and Takeovers) Regulations 2011, SEBI (Prohibition of Insider Trading) Regulations 2015, SEBI (Alternative Investment Funds) Regulations 2012, SEBI (Portfolio Managers) Regulations 2020, SEBI (Research Analysts) Regulations 2014, SEBI (Buy-Back of Securities) Regulations 2018, SEBI (Delisting of Equity Shares) Regulations 2021, SEBI (Mutual Funds) Regulations 1996, SEBI (Depositories and Participants) Regulations 2018, SEBI (Merchant Bankers) Regulations 1992, recent SEBI circulars (2024--2026).

\textbf{RBI Documents:} RBI Monetary Policy Statements (2024--2026), RBI Master Directions on Unique Identifiers in Financial Markets, RBI (Small Finance Banks---Prudential Norms on Declaration of Dividend) Directions 2026, Government Securities auction notifications (2025--2026), State Government securities auction press releases, RBI Weekly Statistical Supplement extracts, RBI circulars on KYC/AML compliance.

\section{System Prompt Template}
\label{app:prompt}

\begin{quote}
\noindent You are an expert in Indian financial regulation and policy. Answer questions using ONLY the provided context passage. Do not use any external knowledge. Be concise and precise. Give only the answer --- no preamble.
\end{quote}

\section{CON Task: Majority-Class Baseline and Balanced Accuracy}
\label{app:con}

The contradiction detection task (CON) has a class-imbalanced label distribution: 53 of 62 items (85.5\%) carry the gold label `No' (no contradiction exists), yielding a majority-class baseline accuracy of 85.5\%. Table~\ref{tab:c1} reports raw CON accuracy, margin over that baseline, and balanced accuracy (mean of sensitivity and specificity) computed from each model's per-item predictions. All models except Gemma 3 4B and Mistral-7B exceed the majority-class baseline. Balanced accuracy preserves the raw-accuracy ordering; notably, most models detect the nine true contradictions perfectly (sensitivity 100\%) and differ mainly in false-positive rate on the 53 `No' items.

\begin{table}[t]
\centering\small
\begin{tabular}{lrrr}
\toprule
\textbf{Model} & \textbf{CON \%} & \textbf{vs. Base} & \textbf{Bal. Acc.} \\
\midrule
Gemini 2.5 Flash & 88.7 & +3.2 & 88.8 \\
Qwen3-32B & 90.3 & +4.8 & 94.3 \\
LLaMA-3.3-70B & 95.2 & +9.7 & 97.2 \\
Llama 4 Scout 17B & 98.4 & +12.9 & 99.1 \\
Kimi K2 & 91.9 & +6.4 & 95.3 \\
LLaMA-3-8B & 93.5 & +8.0 & 91.6 \\
GPT-OSS 120B & 95.2 & +9.7 & 97.2 \\
GPT-OSS 20B & 95.2 & +9.7 & 97.2 \\
Gemini 2.5 Pro & 93.5 & +8.0 & 91.6 \\
Mistral-7B & 80.6 & $-$4.9 & 88.7 \\
DeepSeek-R1-Distill-Llama-70B & 96.8 & +11.3 & 98.1 \\
Gemma 3 4B & 79.0 & $-$6.5 & 87.7 \\
\midrule
Majority baseline & 85.5 & --- & 50.0 \\
\bottomrule
\end{tabular}
\caption{CON task raw accuracy, margin over the majority-class baseline (85.5\%), and balanced accuracy (mean of sensitivity and specificity over 9 `Yes' / 53 `No' gold labels).}
\label{tab:c1}
\end{table}

\section{Scoring Pipeline Validation}
\label{app:judge}

The automated scoring pipeline (Section~\ref{sec:scoring}) uses four stages: exact match, fuzzy token matching (RapidFuzz \texttt{token\_set\_ratio} $\geq$ 0.72), numerical extraction match, and Yes/No match for contradiction detection. While rigorous, the pipeline penalizes answers that are semantically correct but formatted differently from the reference answer---a systematic limitation of string-based matching for open-ended regulatory QA.

\subsection{Judge Rubric}

Both judges used in this paper---the Gemini 2.5 Flash pilot described below and the phi4-mini cross-model judge of Section~\ref{sec:judge}---were given the identical rubric, at temperature 0:

\begin{quote}\small\noindent
Determine if the model prediction is CORRECT.\\
Rules:\\
--- CORRECT if the final value or fact matches, even if: different formatting (currency symbols, comma separators, units), extra calculation steps shown before the final answer, paraphrasing of the same regulatory rule, or rounding within 1\%.\\
--- INCORRECT only if: wrong number, wrong threshold, wrong date, wrong entity, or contradicts the reference.\\
Reply with exactly: CORRECT or INCORRECT, plus a one-sentence reason.
\end{quote}

\subsection{Judge-Tolerance Sensitivity}
\label{app:judge-tolerance}

The rubric above accepts rounding within 1\%. Because this paper's central claim is that
scoring-rule choices are themselves consequential, that tolerance is exactly the kind of parameter
the claim should be tested against rather than left as an undefended default. We re-ran the
full-coverage phi4-mini judge (Section~\ref{sec:judge}) at two additional tolerances---exact numeric
match (no rounding clause) and 0.1\%---under otherwise identical conditions (same model, temperature,
context sizing, and full REG/NUM/TMP coverage), and compare all three against the 1\% run used
throughout the paper.

\begin{table*}[t]
\centering\small
\begin{tabular}{lrrrrrr}
\toprule
 & \multicolumn{2}{c}{\textbf{Exact}} & \multicolumn{2}{c}{\textbf{0.1\%}} & \multicolumn{2}{c}{\textbf{1\% (paper)}} \\
\textbf{Model} & \textbf{Judge-only} & \textbf{Rank} & \textbf{Judge-only} & \textbf{Rank} & \textbf{Judge-only} & \textbf{Rank} \\
\midrule
Gemini 2.5 Pro & 95.57 & 2 & 95.07 & 3 & 94.83 & 1 \\
Llama 4 Scout 17B & 95.57 & 2 & 95.07 & 3 & 94.33 & 2 \\
GPT-OSS 120B & 95.57 & 2 & 95.32 & 2 & 93.84 & 3 \\
LLaMA-3-8B & 94.09 & 4 & 92.36 & 8 & 93.60 & 4 \\
Kimi K2 & 95.57 & 2 & 95.81 & 1 & 93.35 & 5 \\
GPT-OSS 20B & 96.80 & 1 & 95.81 & 1 & 93.10 & 6 \\
Qwen3-32B & 92.86 & 5 & 93.10 & 6 & 92.36 & 7 \\
\textbf{DeepSeek-R1-Distill} & \textbf{95.57} & \textbf{2} & \textbf{94.58} & \textbf{4} & \textbf{92.36} & \textbf{7} \\
Mistral-7B & 89.41 & 6 & 88.92 & 9 & 91.38 & 8 \\
Gemini 2.5 Flash & 92.86 & 5 & 92.86 & 7 & 89.66 & 9 \\
LLaMA-3.3-70B & 94.33 & 3 & 93.60 & 5 & 88.67 & 10 \\
Gemma 3 4B & 87.68 & 7 & 87.68 & 10 & 87.19 & 11 \\
\bottomrule
\end{tabular}

\caption{Judge-only accuracy and rank at three rounding tolerances, all twelve models, sorted by the
1\% (paper) rank. Same phi4-mini judge and full REG/NUM/TMP coverage at every tolerance; only the
rubric's rounding clause differs.}
\label{tab:judgetolerance}
\end{table*}

Two findings, in different directions. First, the paper's central claim is robust: strict and
judge-only show no significant rank correspondence at any tolerance (Spearman $\rho = -0.32$,
$-0.13$, $-0.27$ at exact, 0.1\%, and 1\% respectively, all $p > 0.3$), and the judge-only spread
stays within a similar 7.6--9.1 point band throughout. Second, individual models' judge-only
\emph{ranks} are more tolerance-sensitive than the headline numbers suggest: seven of twelve models
move 4 or more ranks across the three tolerances tested, most sharply LLaMA-3.3-70B (rank 3 to 10, 7
positions) and GPT-OSS 20B (rank 1 to 6, 5 positions). DeepSeek-R1-Distill itself moves from
judge-only rank 2 (exact) to 4 (0.1\%) to 7 (1\%)---a real, disclosed dependency on tolerance,
though it never approaches its strict rank of 12 at any tolerance tested. Under judge-augmented
scoring specifically, DeepSeek-R1-Distill is 1st outright at exact tolerance (98.77\%, no tie),
tied for 2nd at 0.1\% (98.28\%, with GPT-OSS 20B and LLaMA-3.3-70B), and tied for 1st at 1\% (98.03\%,
with LLaMA-3.3-70B---the paper's headline finding): rank 1 or 2 at every tolerance tested, the more
stable of its two headline results.

We read this as reinforcing rather than undermining the paper's thesis, not as a reason to prefer
one tolerance over another. The finding is not that 1\% was a poor choice---judge-only's defining
property, no positive correspondence with strict, holds regardless of which of the three tolerances
is used---it is that scoring-rule sensitivity extends one level deeper than the strict-versus-judge
comparison alone: even within a fixed judge model and prompt, a single numeric rubric parameter
moves individual model ranks by several positions. A benchmark that reports judge-only accuracy at
one tolerance without stating it as a choice among defensible alternatives understates how much of
the reported ranking is attributable to that specific choice.

\subsection{Gemini Pilot}

Before committing to a full-coverage second judge, we piloted semantic re-scoring with Gemini 2.5 Flash---one of the twelve evaluated models---over items strict scoring marked incorrect across NUM, REG, and TMP. The strict pipeline flagged 935 such predictions; 874 were judged, and the remaining 61 (58 empty responses, 3 API failures) were retained as incorrect without review, since no semantic judgment can rescue an empty prediction. CON items use exact Yes/No matching and were not part of this pilot.

Judge reliability was estimated on a stratified 28-item manual audit (20 NUM, 5 REG, 3 TMP). Manual review confirmed 25 of 28 verdicts (89.3\%) as correct. The three judge errors all involved multi-step numerical problems where the model showed correct intermediate reasoning but reached a wrong final answer (NUM\_074: predicted Rs.~19,000 crore vs.\ reference Rs.~33,000 crore; NUM\_103: concluded wrong direction for a threshold comparison; NUM\_136: applied wrong regulatory fee slab)---the judge's tendency to over-credit partial reasoning chains on NUM items, referenced in Section~\ref{sec:discussion}. Across the 874 judged predictions, Gemini reclassified 685 from incorrect to correct, an overall flip rate of 78.4\% (REG: 261/327 = 79.8\%; NUM: 272/348 = 78.2\%; TMP: 152/199 = 76.4\%).

Because Gemini is itself one of the evaluated models and this pilot covered only strict failures---so it could characterise strict-to-judge overturns but not the judge's rejection rate of strict-correct predictions---these results motivated, rather than substitute for, the full-coverage cross-model judge (phi4-mini, sharing no model family with any evaluated system, reviewing all 4{,}128 REG/NUM/TMP predictions including strict-correct ones) whose results are reported in Section~\ref{sec:judge} and Table~\ref{tab:judgereliability}. Item-level Gemini verdicts remain available at \texttt{evaluation/results\_judged/} in the repository; the judge-only and judge-augmented accuracies used throughout this paper are computed from phi4-mini's full-coverage pass (\texttt{evaluation/results\_judged\_phi4/}), not from this pilot.

\subsection{phi4-mini Pilot Validation}

Before committing to the full cross-model pass, we validated phi4-mini on two small, targeted
samples: it agreed with all eight items Gemini 2.5 Flash itself answered correctly under strict
scoring (a sanity check, not evidence of discrimination), and on eight DeepSeek-R1-Distill items
marked incorrect under strict scoring, flipped seven to correct---each a genuine format
mismatch---while correctly holding the eighth, where the predicted figure simply did not match the
reference. The pilot showed that phi4-mini both overturned strict failures and retained cases where
the substantive answer remained incorrect.

\subsection{Root Causes}

Reading a sample of overturned items during the Gemini pilot surfaces three recurring surface-form patterns. This is a qualitative description of what these items look like, not a quantified account of what predicts reclassification in general---Section~\ref{sec:regime} tests output length directly and finds no association that survives adjustment for task and model composition, so none of the three should be read as an established mechanism. (1) \textit{Numeric format differences} --- models frequently use digits (`50\%', `365 days') while references use written numbers (`fifty per cent', `three hundred and sixty-five days'). (2) \textit{Explanatory prefixes} --- models prepend calculation steps or regulatory citations before stating the final value. (3) \textit{Abbreviated answers} --- models give a shorter but correct version of the reference (e.g., ref = `Six months after completion of the open offer'; pred = `Six months').

\subsection{Logging Truncation}
\label{app:truncation}

Predictions are character-capped at write time by the evaluation harness, at a limit that varies by
which script produced the row (200, 300, or 500 characters). 554 of the 4{,}128 judged REG/NUM/TMP
rows (13.4\%) sit exactly at their model's cap, ranging from 31.1\% (Gemini 2.5 Flash, 200-char cap)
to 0.3\% (LLaMA-3-8B and Mistral-7B, both repaired to a higher effective cap during the truncation
fix described below). The judge previously imposed a second, undisclosed layer of truncation on its
own inputs (question capped at 600 characters, reference at 300, prediction at 500), independent of
the logging cap above; this has been removed. \texttt{scripts/judge\_phi4\_crossmodel.py} now sizes
Ollama's context window (\texttt{num\_ctx}) per run from the longest question+reference+prediction
triple in the corpus being judged, with a 25\% safety margin over the estimated token count plus the
150-token output budget---2{,}048 tokens for the released judged data, logged in
\texttt{evaluation/judge\_run\_manifest.json} together with an assertion, checked per item, that every
prompt fits; zero of the 4{,}128 judged rows were flagged as not fitting. The 449 rows whose reference
answer or prediction had been cut by the old 300/500-character judge-input caps (concentrated on
Qwen3-32B, Llama 4 Scout 17B, LLaMA-3-8B, Mistral-7B, and Gemini 2.5 Pro's judged files) were
re-judged under the fixed pipeline: 52 verdicts changed (19 strict-wrong items newly judged correct,
33 newly judged incorrect), a mixed-direction shift consistent with removing a text-availability
artifact rather than an infrastructure failure. No response includes a provider-reported finish-reason,
stop-reason, or token-usage field anywhere in the released data, so generation-level truncation---as
opposed to the logging truncation quantified here---cannot be measured from the original release at
all; the matched-budget re-run (Section~\ref{sec:budget}) closes this gap directly, with full,
untruncated responses and finish-reason logging for every model reached.

Restricting to the 190 of 344 REG/NUM/TMP items where every one of the twelve models' predictions is
strictly below its own cap---so no model's prediction on that item could possibly be a truncation
artifact---DeepSeek-R1-Distill-Llama-70B remains last of twelve under strict scoring (70.0\%) and
reaches the top tier under the judge-augmented sensitivity regime (98.4\%, 5-way tie for the top
rank). The qualitative pattern this paper reports is not an artifact of logging truncation, though
the exact ranks and margins on the truncation-free subset differ modestly from the full-sample
figures, as expected with a smaller item count.

\subsection{Human Adjudication Protocol}

To check both judges against a human reader, we built a 238-item adjudication sheet (\texttt{annotation/judge\_audit/adjudication\_sheet.csv}, \texttt{scripts/build\_adjudication\_control\_sample.py}): the 174 REG/NUM/TMP items where Gemini and phi4-mini disagree---the only cases informative about which judge is right---plus a stratified control sample of 64 items (8 per stratum, fixed random seed) spanning cases where the available methods \emph{agree}: \texttt{\{REG,NUM,TMP\}\_strict\_incorrect\_judges\_agree} (both judges concur an item is wrong), \texttt{\{REG,NUM,TMP\}\_strict\_correct\_phi4\_agrees} (phi4-mini does not dispute a strict-correct item; no Gemini verdict exists for these, since Gemini only ever judged strict failures), and two CON strata (\texttt{CON\_strict\_correct\_no\_judge}, \texttt{CON\_strict\_incorrect\_no\_judge}) that spot-check the exact-match scorer directly, since CON is never judge-reviewed by either model. This sample is evidence about specific, informative cases, not a random draw sized to estimate a population-level judge error rate, and Section~\ref{sec:judge} reports it as such.

\subsection{Human Adjudication Results}

All 238 items were adjudicated by the author, scored against the judge verdicts as they stood at the time of adjudication (Section~\ref{sec:judge}). Section~\ref{sec:judge} reports the headline split (89.1\% control-agreement, 56.9\%/43.1\% Gemini/phi4-mini on disputed items); per stratum, the control-agreement rate ranged from 75.0\% (\texttt{NUM\_strict\_incorrect\_judges\_agree}, \texttt{TMP\_strict\_correct\_phi4\_agrees}) to 100\% (both CON strata and \texttt{REG\_strict\_correct\_phi4\_agrees}), with no stratum below 75\%---agreement between methods tracked human judgment reasonably well across every category sampled, not only on average. On the 174 disagreement items the human/Gemini--human/phi4-mini split is widest on numerical reasoning (Gemini matched 44 of 68, 64.7\%, against phi4-mini's 24, 35.3\%) and narrows to near-even on regulatory interpretation (30 of 59 vs.\ 29 of 59) and temporal reasoning (25 of 47 vs.\ 22 of 47). Section~\ref{sec:limitations} reports the DeepSeek-R1-Distill-specific breakdown of the disputed-item sample.

\subsection{Independent (Non-Author) Adjudication}
\label{app:independent-adjudication}

The adjudication above was performed by the author, which leaves open whether it reflects a
personal reading rather than one a different human would reach. To check this, two independent
adjudicators---acquaintances of the author with no role in constructing the benchmark, blind to
model identity, to both judges' verdicts, and to each other's answers throughout---each completed
the full 62-item stratified subsample of the 174-item Gemini/phi4-mini disagreement set (drawn with
a fixed seed, stratified by task, disagreement direction, strict-correctness status, prediction
length, and truncation status; \texttt{scripts/build\_independent\_adjudication\_sample.py}), via a
purpose-built review page presenting one item at a time with no technical interface
(\texttt{scripts/build\_adjudication\_artifact.py}). Agreement is computed directly from both
completed sheets (\texttt{scripts/score\_independent\_adjudication.py}); we report raw percent
agreement and Cohen's $\kappa$ throughout, never treating any of the three human verdicts (author,
or either independent adjudicator) as ground truth over the others.

The two independent adjudicators agree with each other on 51 of 62 items (82.3\%, $\kappa=0.652$)
and with the author's own earlier verdicts at 87.1\% ($\kappa=0.739$) and 82.3\% ($\kappa=0.647$)
respectively---all three pairwise comparisons fall in the ``substantial agreement'' band by
\citet{landiskoch1977} convention, the same convention used for the CON inter-annotator statistic in
Table~\ref{tab:validation}. This does not establish that the author's original adjudication was correct on
any specific item, but it is evidence against the narrower concern that a single self-adjudicating
author's verdicts on this sample are idiosyncratic: two people with no stake in the benchmark and no
knowledge of the author's own verdicts landed on the same reading at a rate comparable to expert
human agreement rates reported for similar tasks. Agreement is not uniform across tasks: regulatory
interpretation and numerical reasoning items show strong three-way agreement (76--100\% pairwise
across all three adjudicators), while temporal reasoning items are markedly harder to adjudicate
consistently (62.5\% pairwise agreement between the two independent adjudicators, and between each
and the author)---consistent with temporal-reasoning items already being the ones this paper's
error-taxonomy analysis (Appendix~\ref{sec:errortax-sub}) finds hardest for the judge to clear
cleanly, now shown to be hard for independent human readers as well, not only for the automated
pipeline.

Against the two judges, both independent adjudicators land close to the author's own split and to
each other: Gemini agreement is 51.6\% (Adjudicator A) and 53.2\% (Adjudicator B), phi4-mini
agreement is 48.4\% and 46.8\% respectively, against the author's own 51.6\%/48.4\% on the identical
62-item subsample.
A majority vote across all three human verdicts reproduces the same 51.6\%/48.4\% split exactly.
Three independent readings therefore converge on the same qualitative picture as the author's
single-handed one: on genuinely disputed items, human readers side with Gemini over phi4-mini only
slightly more often than the reverse, and neither judge is right often enough on these cases to
stand in for ground truth. We report this as independent replication of that specific finding, not
as validation of either judge---the wording distinction matters, since ``the judge was validated by
humans'' is a stronger and different claim than ``independent humans replicate the same disagreement
pattern the author found alone,'' and only the latter is supported here.

\section{Evaluated Models}
\label{app:models}

\begin{table*}[t]
\centering\footnotesize
\resizebox{\textwidth}{!}{%
\begin{tabular}{lllrr}
\toprule
\textbf{Model} & \textbf{Provider} & \textbf{API model / Ollama tag} & \textbf{Params} & \textbf{Budget} \\
\midrule
Gemini 2.5 Flash & Google (AI Studio API) & \texttt{gemini-2.5-flash} & --- & 200 \\
Gemini 2.5 Pro & Google (Vertex AI) & \texttt{gemini-2.5-pro} & --- & none\textsuperscript{a} \\
Qwen3-32B & Groq API & \texttt{qwen/qwen3-32b} & 32B & 1024 \\
LLaMA-3.3-70B & Groq API & \texttt{llama-3.3-70b-versatile} & 70B & 200 \\
Llama 4 Scout 17B & Groq API & \texttt{meta-llama/llama-4-scout-17b-16e-instruct} & 17B & 1024 \\
Kimi K2 & Groq API & \texttt{moonshotai/kimi-k2-instruct} & 1T total, 32B active (MoE) & 512 \\
LLaMA-3-8B & Ollama, local & \texttt{llama3}\textsuperscript{b} & 8B & 300 \\
GPT-OSS 120B\textdaggerdbl{} & Groq API & \texttt{openai/gpt-oss-120b} & 120B & 512 \\
GPT-OSS 20B\textdaggerdbl{} & Groq API & \texttt{openai/gpt-oss-20b} & 20B & 512 \\
Mistral-7B & Ollama, local & \texttt{mistral}\textsuperscript{b} & 7B & 300 \\
DeepSeek-R1-Distill-Llama-70B & OpenRouter & \texttt{deepseek/deepseek-r1-distill-llama-70b} & 70B & 2048 \\
Gemma 3 4B & Ollama, local & \texttt{gemma3:4b} & 4B & 512 \\
\bottomrule
\end{tabular}%
}
\caption{Models evaluated in this study, with the exact API model identifier or Ollama tag passed
to each provider and the generation completion budget (tokens) set for it, both read directly from
the evaluation harness (\texttt{scripts/evaluate*.py}). Budgets were set per script for cost reasons
and are \textbf{not} uniform across models; Section~\ref{sec:models} discusses this as a covariate.
\textdaggerdbl{}OpenAI open-weight models accessed via the Groq inference API. Local models ran on
an Intel i7-13650HX CPU with an NVIDIA RTX 4060 GPU (8~GB VRAM) via the Ollama REST API, temperature
0.0. \textsuperscript{a}Vertex AI's default for this model; the harness set no explicit cap.
\textsuperscript{b}The harness calls the bare Ollama tags \texttt{llama3} and \texttt{mistral}, which
resolved to the \texttt{llama3:8b} and \texttt{mistral:7b} weights via a local \texttt{:latest}
alias; the released result files separately record these as
\texttt{meta-llama/Meta-Llama-3-8B-Instruct} and \texttt{mistralai/Mistral-7B-Instruct-v0.3}, a
manually-assigned display label for the same weights, not a different model. Local weight identity
is pinned by Ollama's own content digest (\texttt{ollama show}/\texttt{/api/tags}, SHA-256 of the
GGUF blob, re-confirmed 2026-09-03): \texttt{llama3:latest} = \texttt{365c0bd3c000}\ldots{} (Q4\_0,
8.0B params, 8192-token context); \texttt{mistral:latest} = \texttt{6577803aa9a0}\ldots{} (Q4\_K\_M,
7.2B params, 32768-token context); \texttt{gemma3:4b} = \texttt{a2af6cc3eb7f}\ldots{} (Q4\_K\_M,
4.3B params). Full 64-character digests are released with the evaluation code.}
\label{tab:models}
\end{table*}

\subsection{Matched-Budget Results}
\label{app:matched-budget-results}

Section~\ref{sec:models} disclosed that the original evaluation used per-model completion budgets
that were not uniform (200--2048 tokens, chosen per provider for cost reasons) and that write-time
logging separately capped stored predictions at 200--500 characters regardless of what the model
generated (Appendix~\ref{app:truncation}). Both are operational choices, not properties of the
models, and both can depress a model's strict score independently of its underlying answer. We
re-ran the 11 of 12 models reachable under current API credentials at a single uniform completion
budget of 512 tokens---selected by pilot (below)---logging the complete prediction plus provider
finish-reason and completion-token count for every item, neither of which the original release
records for any model. Gemini 2.5 Pro is permanently unreachable on any fresh API key (below gives
the exact platform error).

\begin{table}[t]
\centering\small
\begin{tabular}{lrrrrr}
\toprule
Model & Orig.\ budget & Original & Matched (512) & $\Delta$pp & Trunc.\ @512 \\
\midrule
LLaMA-3-8B & 300 & 78.08 & 82.27 & +4.19 & 0.0\% \\
Mistral-7B & 300 & 75.86 & 77.83 & +1.97 & 0.0\% \\
Gemma 3 4B & 512 & 78.82 & 78.82 & +0.00 & 0.0\% \\
GPT-OSS 120B & 512 & 77.09 & 76.85 & -0.24 & 1.7\% \\
GPT-OSS 20B & 512 & 76.85 & 74.88 & -1.97 & 5.2\% \\
DeepSeek-R1-Distill & 2048 & 75.12 & 85.71 & \textbf{+10.59} & 1.2\% \\
LLaMA-3.3-70B & 200 & 83.74 & 84.73 & +0.99 & 0.2\% \\
Llama 4 Scout 17B & 1024 & 83.25 & 80.79 & -2.46 & 0.0\% \\
Kimi K2 & 512 & 81.53 & 81.28 & -0.25 & 5.9\% \\
Qwen3-32B & 1024 & 85.47 & 83.99 & -1.48 & 0.0\% \\
Gemini 2.5 Flash & 200 & 89.66 & 85.96 & -3.70 & 0.0\% \\
\bottomrule
\end{tabular}

\caption{Strict accuracy under the archived original release (per-model budget, original provider)
versus a re-run at a uniform 512-token budget, same four-stage scorer, same items. The two columns
differ in budget, logging cap, and---for five models---transport (four moved from Groq to
OpenRouter; Gemini 2.5 Flash moved from direct Google AI Studio access to the same model routed
through OpenRouter; text below), so $\Delta$pp is a joint evaluation-configuration effect, not an
isolated budget effect; it is bolded where $|\Delta| \geq 5$. ``Trunc.\ @512'' is the fraction of
matched-run predictions that hit the 512-token cap (\texttt{finish\_reason=length}). Original
completion budgets and truncation rates: Appendix~\ref{app:truncation}.}
\label{tab:matchedbudget}
\end{table}

This re-run changes several things relative to the original release at once---completion budget, the
removal of the write-time logging cap, and, for five of the eleven models, a change of transport
(four moved from Groq to OpenRouter, forced by Groq's deprecation of those model endpoints; Gemini
2.5 Flash moved from direct Google AI Studio access to the same model routed through OpenRouter's
Google AI Studio integration, a transport change made to reach a usable quota after the direct
route stalled for hours; details below)---so a per-model $\Delta$ here is their joint effect, not an
isolated budget effect; the individual contributions cannot be separated post hoc from released data
that never recorded generation-level truncation in the first place. We therefore treat this as a
test of joint sensitivity to evaluation configuration, not a controlled budget experiment. Ten of the
eleven models shift by at most 4.2 points in either direction ($-3.70$ to $+4.19$pp), consistent with
a modest, mixed-sign operational effect on top of measurement noise---and, as the following
subsection details, the two largest positive shifts among those ten belong specifically to the two
models whose serving configuration did not change at all (the local Ollama models), while Gemini 2.5
Flash---originally the highest-truncation model in the study at 31.1\% and now also
transport-changed---shows the largest negative shift among the ten, consistent with serving/transport
variance contributing at least as much as budget itself. DeepSeek-R1-Distill is an outlier at
$+10.59$pp; we report it without endorsing a single explanation, since it does not positively
correlate with each model's original write-time-cap rate (Pearson $r = -0.77$ across the other ten,
if anything in the opposite direction from a truncation-recovery story) and at least one competing
explanation, unrelated to budget, remains live (the following subsection gives both candidates and
the evidence for each in full). What the re-run supports without qualification is the general point:
strict accuracy is sensitive to the combined evaluation configuration---completion budget, logging
configuration, and serving provider jointly, not any one in isolation---sometimes by several points.
Further evidence, alongside the scoring-regime result, that a single leaderboard number is a property
of the measurement protocol as much as of the model.

\subsection{Matched-Budget Re-run: Provider and Model Identifiers}
\label{app:matched-budget-ids}

Four of the original Groq-hosted models (LLaMA-3.3-70B, Llama 4 Scout 17B, Qwen3-32B, Kimi K2) were
confirmed removed from Groq's own model catalogue as of 2026-09-03 (a live \texttt{models.list()}
call against Groq returns only the two GPT-OSS models from the original six); their open-weight
checkpoints were re-run via OpenRouter instead, under exact identifiers verified live against
OpenRouter's models endpoint: \texttt{meta-llama/llama-3.3-70b-instruct} (Meta: Llama 3.3 70B
Instruct), \texttt{meta-llama/llama-4-scout} (Meta: Llama 4 Scout), \texttt{qwen/qwen3-32b} (Qwen:
Qwen3 32B), \texttt{moonshotai/kimi-k2} (MoonshotAI: Kimi K2, the \texttt{0711} dated snapshot ---
the original Groq-served checkpoint's exact build date is not recoverable, since Groq no longer
serves it to check against). Qwen3-32B's OpenRouter route (Nebius) answers in extended-thinking mode
by default, consuming budget on invisible reasoning before any visible answer; every Qwen3-32B call
in the matched-budget re-run appends the model's own documented \texttt{/no\_think} directive to
suppress this, matching the direct-answer, no-preamble system prompt used throughout. DeepSeek-R1-
Distill-Llama-70B was already OpenRouter-hosted in the original run and uses the same identifier,
\texttt{deepseek/deepseek-r1-distill-llama-70b}, unchanged. Gemini 2.5 Pro is excluded from the
matched-budget re-run: a live call against \texttt{gemini-2.5-pro} on a fresh API key returns
\texttt{404 NOT\_FOUND}, ``This model \ldots{} is no longer available to new users'' --- a platform
access restriction, not a credentials problem, and not something a different model generation (e.g.
\texttt{gemini-3.1-pro-preview}) can substitute for without re-running a different model than the
one this paper reports on.

Gemini 2.5 Flash's direct Google AI Studio route throttled after roughly 20 requests regardless of
request spacing (confirmed at 8s and 15s spacing, via the exact
\texttt{generate\_content\_free\_tier\_requests} quota-exceeded message, well below the model's
documented rate limit, consistent with a new-project ramp-up restriction), and did not clear across
several hours of intermittent checking. We re-ran it instead through OpenRouter's routing to the
same model (\texttt{google/gemini-2.5-flash}), with the request explicitly restricted to Google AI
Studio as the serving provider (\texttt{provider.only=["google-ai-studio"]}) at its discounted
Flex service tier (\$0.15/M prompt, \$1.25/M completion tokens, verified live against OpenRouter's
own endpoint-pricing listing before running), a price cap set to prevent any silent fallback to a
costlier tier, and \texttt{reasoning.max\_tokens=0} to disable Gemini's hidden-thinking mode (the
same failure class that required Qwen3-32B's \texttt{/no\_think} directive above). A mandatory cost
probe (8 items, then a 12-item sample stratified across all four task types) confirmed zero
reasoning tokens on every item and projected a full-406-item cost of roughly \$0.02 before the full
run was authorized. All 406 items completed with zero errors at a total cost of \$0.0243, confirmed
served entirely by Google AI Studio at the Flex tier throughout
(\texttt{evaluation/gemini\_flash\_openrouter\_manifest.json}). This is a transport change, not a
model substitution: the same Gemini 2.5 Flash weights, reached through a different request path
after the direct route proved unusable within a practical time budget.

\subsection{Budget Selection and the DeepSeek-R1-Distill Outlier}
\label{app:budget-outlier}

The matched budget (512 tokens) was chosen from a pilot across three candidates (512, 1024, 2048) on
a stratified 24-item subset (8 each of REG/NUM/TMP) across all reachable models
(\texttt{scripts/matched\_budget\_pilot.py}): mean truncation across models was 1.1\% at 512, 0.4\% at
1024, and 0.0\% at 2048---raising the budget past 512 buys a reduction of about one point in an
already-negligible rate, so 512 was taken as the lowest budget that does not materially change
truncation relative to the higher candidates.

DeepSeek-R1-Distill's $+10.59$pp delta (Table~\ref{tab:matchedbudget}) is the one result in this
re-run we could not fully explain and report as such rather than resolving it into a clean story. Two
candidate contributors are on the table, neither ruled in or out. First, logging truncation:
DeepSeek's original release had 19.5\% of its REG/NUM/TMP rows sitting at the old 300-character
write-time cap (Appendix~\ref{app:truncation}), the highest rate of any model with a comparably small
delta, so removing that cap plausibly recovers some of the swing---but a Pearson correlation between
each model's original at-cap rate and its matched-budget delta, computed across the other ten
models (\texttt{scripts/matched\_budget\_correlation.py}), is $r = -0.77$: models with a
\emph{higher} original truncation rate tend to show a \emph{smaller or negative} delta, the opposite
of what a truncation-driven recovery story predicts, and the single most truncated model in the
original release (Gemini 2.5 Flash, 31.1\%) shows the most negative delta of the ten ($-3.70$pp).
The two largest positive deltas among the ten (LLaMA-3-8B, $+4.19$pp; Mistral-7B, $+1.97$pp) belong
to the only two models run locally via Ollama, with unchanged serving between the original run and
the re-run and near-zero original truncation (0.3\% each); every remaining, transport-changed
model---whether it stayed on its original provider (GPT-OSS 120B/20B, Gemma 3 4B), moved to
OpenRouter for the re-run (LLaMA-3.3-70B, Llama 4 Scout 17B, Qwen3-32B, Kimi K2), or moved from
direct access to OpenRouter-routed access to the same provider (Gemini 2.5 Flash)---shows a flat or
negative delta regardless of its original truncation rate. This pattern argues against
logging-truncation removal as the dominant driver of the matched-budget deltas generally, and toward
the second candidate below, though neither explains on its own why DeepSeek specifically moved as
far as it did. Second, output-style or serving variance: a direct probe of the same OpenRouter
endpoint (\texttt{deepseek/deepseek-r1-distill-llama-70b}) made while preparing this analysis
returned \texttt{provider: Novita} with the response's \texttt{content} field empty and a
\texttt{reasoning} field populated instead---the same budget-consumed-by-hidden-reasoning failure
mode Qwen3-32B needed an explicit \texttt{/no\_think} directive to avoid (Section~\ref{app:matched-budget-ids}
above). Manual inspection of matched-run predictions confirms a stylistic shift on several items from
the original's more discursive phrasing (e.g. ``The minimum average operating profit required is
fifteen crore rupees over the preceding three years. \textbackslash n\textbackslash nAnswer: Fifteen
crore rupees.'') to a terser, more exact-match-friendly form (``fifteen crore rupees''), which is
consistent with either explanation and does not distinguish between them. Open-weight checkpoints
served through a third-party aggregator are not guaranteed stable in underlying serving behaviour
between the original evaluation and a re-run weeks later, and we cannot rule out that DeepSeek-R1-
Distill's serving changed in that window independent of the budget we varied. We report the number,
the two candidate explanations, and the evidence against attributing it cleanly to either, rather than
picking the more convenient story.

\section{Task Examples, Pipeline, and Failure Cases}
\label{app:examples}

\begin{figure*}[t]
\centering
\includegraphics[width=0.85\textwidth]{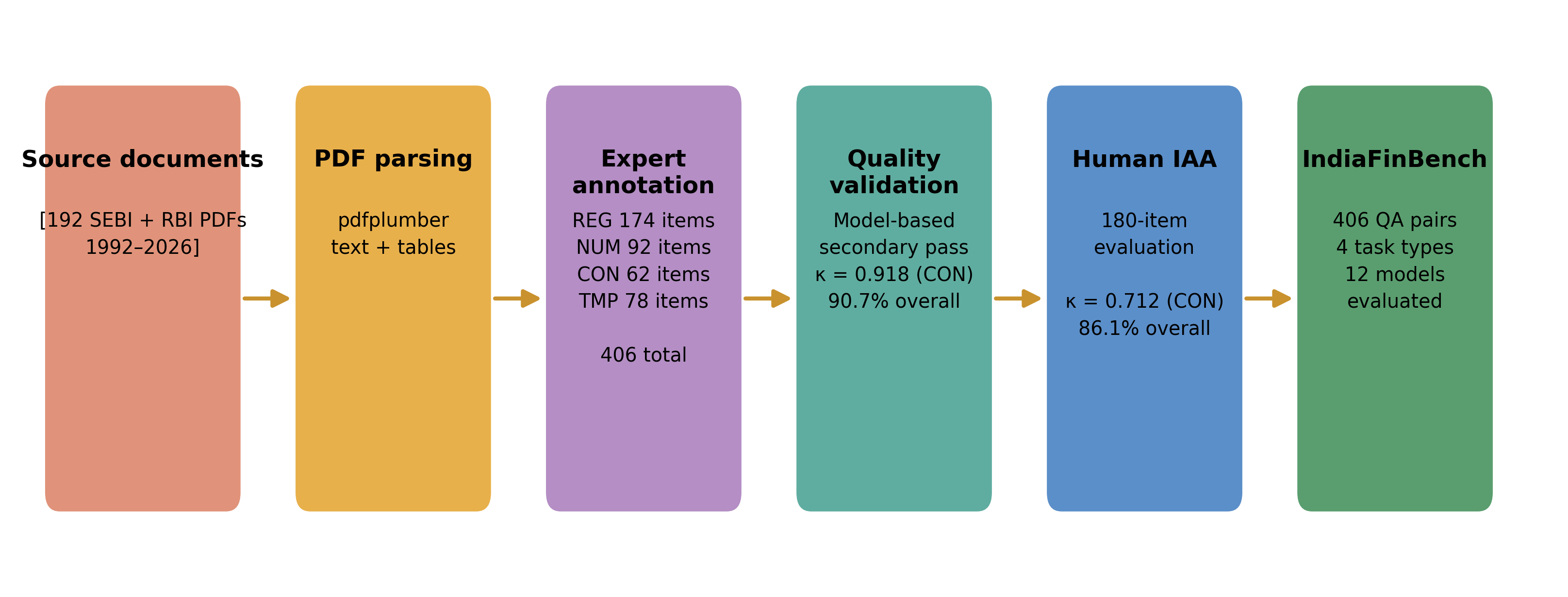}
\caption{IndiaFinBench dataset construction pipeline. Starting from 192 SEBI and RBI PDFs (1992--2026), documents are parsed using pdfplumber and expert-annotated into 406 QA items across four task types (REG: 174, NUM: 92, CON: 62, TMP: 78). Quality is validated through a model-based secondary pass ($\kappa$ = 0.918 on CON; 90.7\% overall) and a 180-item human inter-annotator agreement study ($\kappa$ = 0.712 on CON; 86.1\% overall), yielding a benchmark evaluated on 12 models under zero-shot conditions.}
\label{fig:pipeline}
\end{figure*}

\begin{figure*}[t]
\centering
\includegraphics[width=0.85\textwidth]{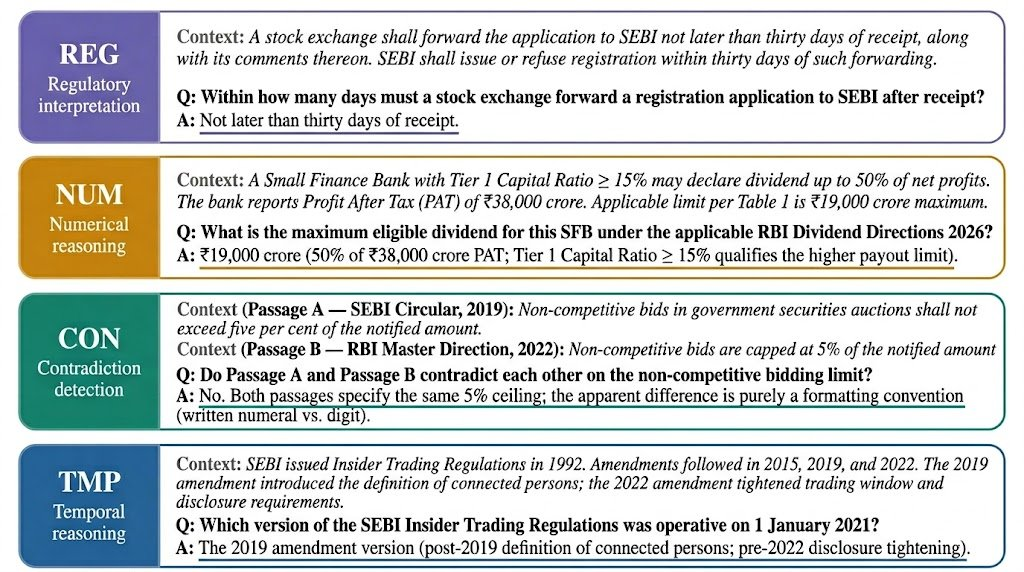}
\caption{Representative examples of all four IndiaFinBench task types. REG (regulatory interpretation): the model extracts the correct compliance deadline from a SEBI passage. NUM (numerical reasoning): the model computes the maximum dividend payout for a Small Finance Bank using RBI Dividend Directions 2026, requiring multi-step arithmetic. CON (contradiction detection): the model determines whether two regulatory passages conflict, identifying that a surface-level phrasing difference (written numeral vs. digit) does not constitute a substantive contradiction. TMP (temporal reasoning): the model identifies the operative version of SEBI Insider Trading Regulations at a specific historical date. Reference answers are underlined.}
\label{fig:taskexamples}
\end{figure*}

\begin{figure}[t]
\centering
\includegraphics[width=\columnwidth]{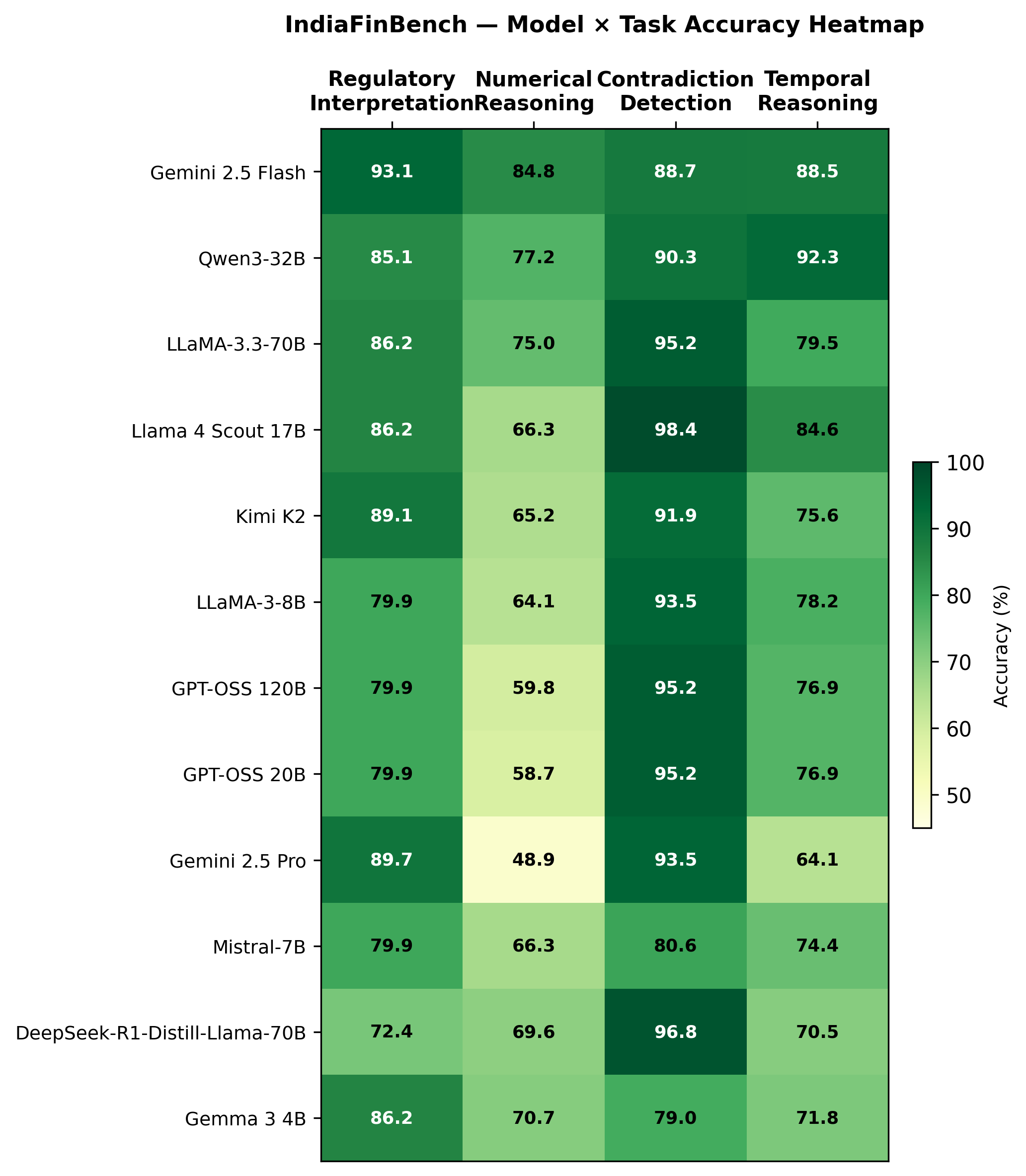}
\caption{Performance heatmap across all twelve models and four task types under strict scoring. Darker cells indicate higher accuracy.}
\label{fig:heatmap}
\end{figure}

Figure~\ref{fig:pipeline} illustrates the dataset construction pipeline. Figure~\ref{fig:taskexamples} shows a representative item for each task type; Figure~\ref{fig:heatmap} visualizes per-task accuracy across all twelve models.

\paragraph{Representative failure examples.}

\textbf{Domain Knowledge Failure (Gemma 3 4B, Temporal Reasoning, item TMP\_039).} Asked whether a December~15, 2026 OTC derivative transaction is subject to Unique Transaction Identifier requirements that take effect January~1, 2027, the model answers ``Yes,'' misapplying the forward-looking effective date to a transaction entered into before it; the reference answer is ``No,'' since the requirements do not yet apply. (Hard temporal-reasoning items map to DKF under the taxonomy of Appendix~\ref{sec:errortax-sub}, since the failure is a substantive misunderstanding of when a rule takes effect, not a surface mismatch.)

\textbf{Numerical Reasoning Failure (GPT-OSS 120B, Numerical Reasoning).} Given an RBI calculation requiring the maximum eligible dividend as a percentage of adjusted PAT, the model correctly identifies the relevant table but applies the wrong conditional threshold, computing the dividend at a higher rate than is warranted by the given capital ratio.

\textbf{Temporal Reasoning Failure (DeepSeek-R1-Distill-Llama-70B, Temporal Reasoning, item TMP\_036).} Asked when the SEBI (Research Analysts) Regulations 2014---published September~1, 2014, effective ninety days later---came into force, the model answers ``November 29, 2014''; the reference date is November 30, 2014, one day later. The model's day-count method is correct; its arithmetic is off by one day.

\textbf{Context Grounding Failure (LLaMA-3.3-70B, Contradiction Detection).} Given two RBI passages specifying the same 5\% non-competitive bidding allocation limit in different phrasing (`five per cent' vs.\ `5\%'), the model incorrectly identifies a contradiction based on surface-level differences rather than recognizing semantic equivalence.

\section{Difficulty Breakdown and Task Profiles}
\label{app:difficulty}

\begin{figure}[t]
\centering
\includegraphics[width=\columnwidth]{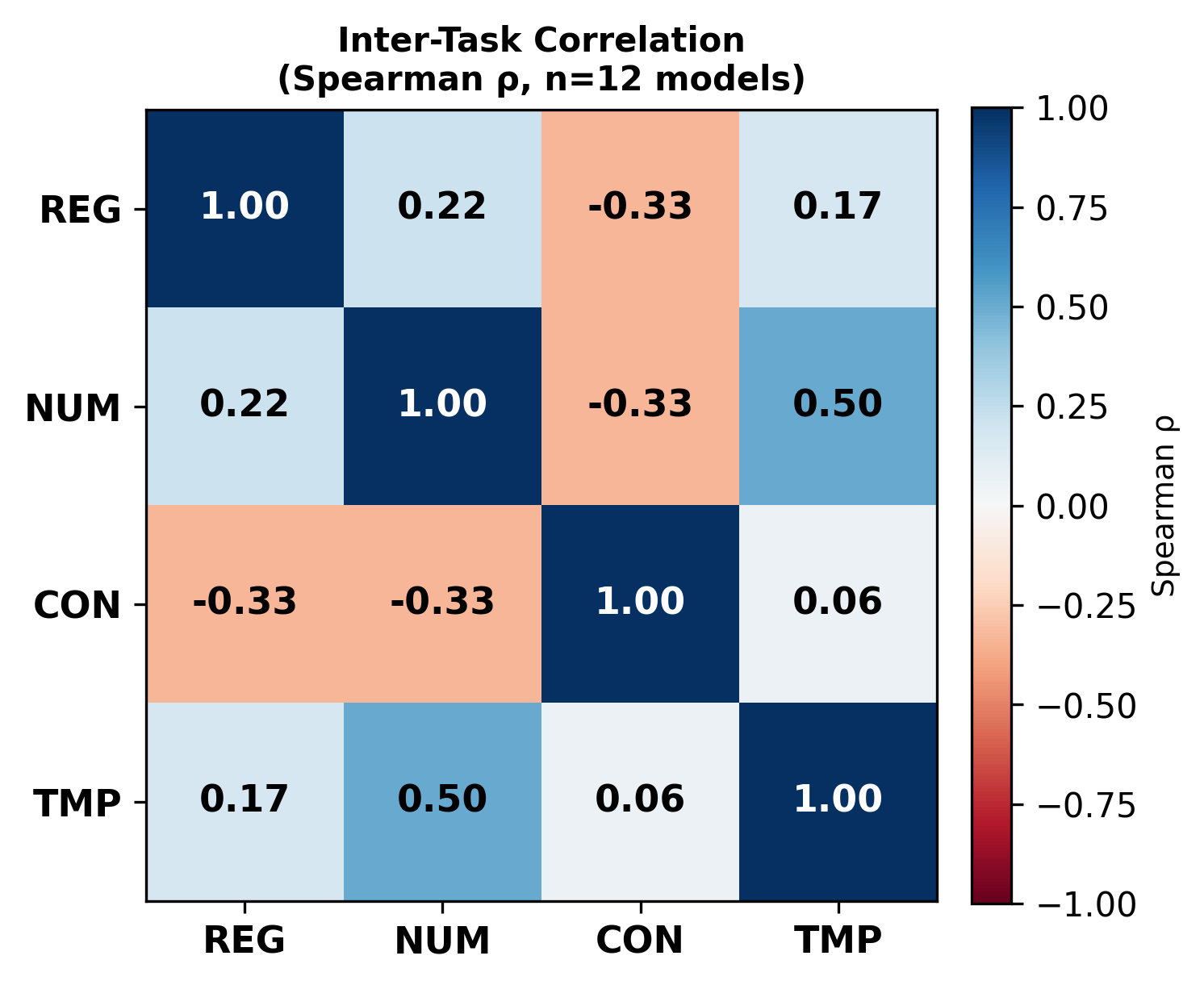}
\caption{Inter-task correlation matrix across the four task types. Values indicate Spearman $\rho$ correlation of per-model accuracy vectors across tasks (n=12 models). With n=12, the critical $\rho$ for significance at p$<$0.05 (two-tailed) is approximately 0.58; no pairwise inter-task correlation reaches this threshold: NUM--TMP is the largest at $\rho$=0.50, REG--CON and NUM--CON are weakly negative ($\rho\approx-0.33$), and the remaining pairs are small and positive. No pairwise inter-task correlation reaches significance in this 12-model sample, suggesting substantial task-specific variation; the sample is too small to support stronger claims about latent capability structure.}
\label{fig:correlation}
\end{figure}

\begin{figure*}[t]
\centering
\includegraphics[width=0.75\textwidth]{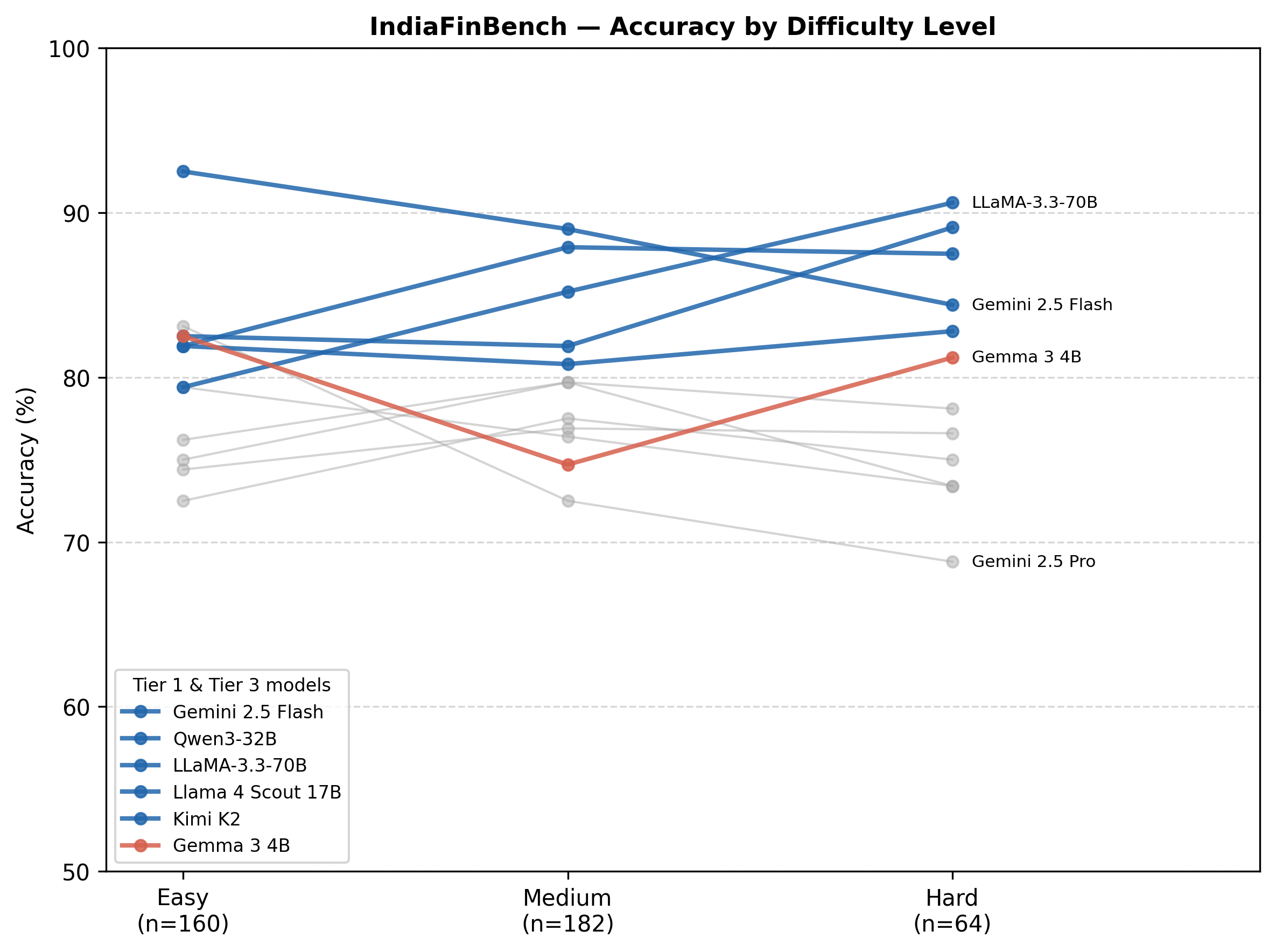}
\caption{Per-model accuracy across difficulty levels (Easy / Medium / Hard). Lines connect the three difficulty conditions for each model.}
\label{fig:difficulty}
\end{figure*}

\begin{figure}[t]
\centering
\includegraphics[width=\columnwidth]{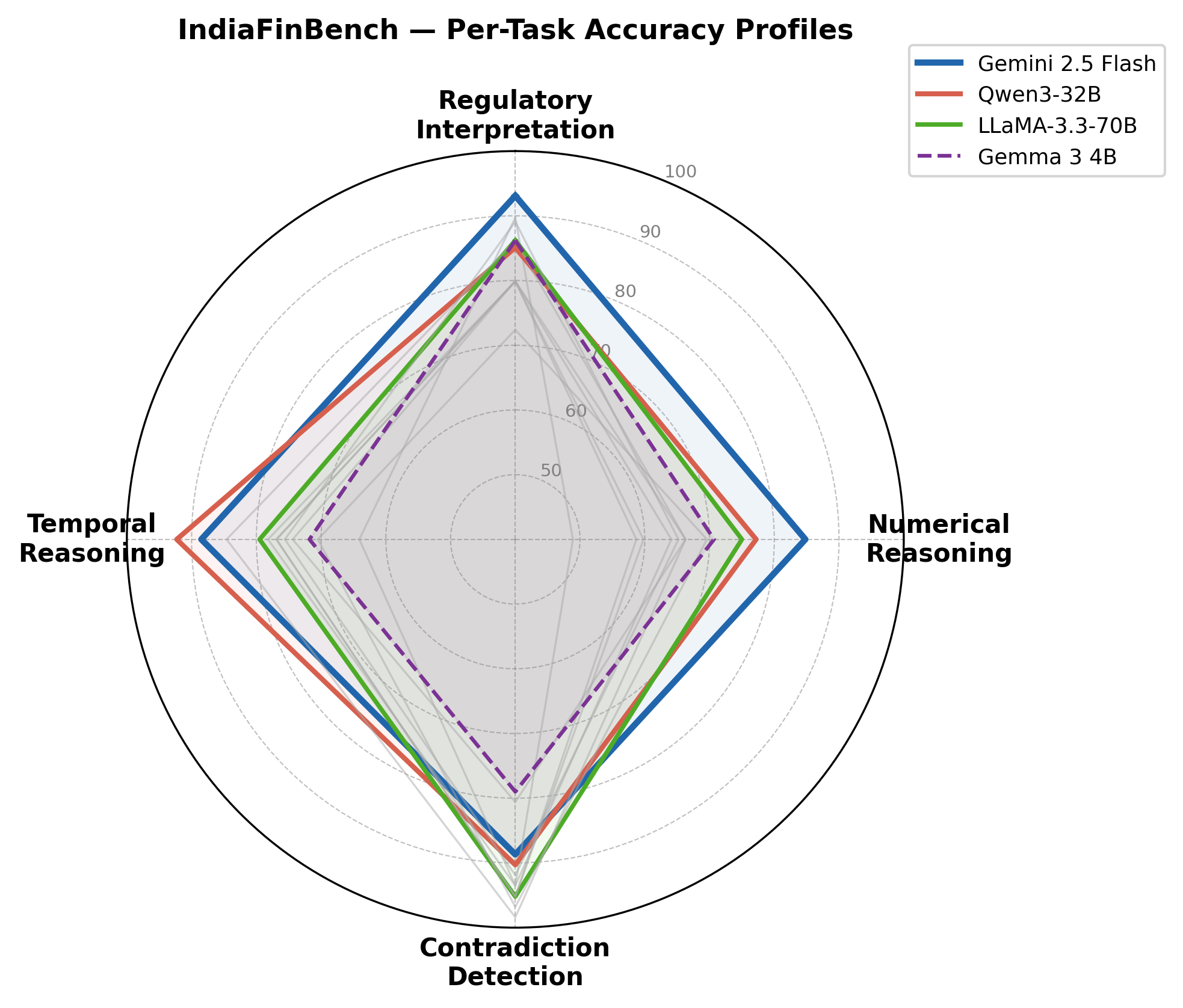}
\caption{Radar chart comparing per-task accuracy profiles across models. Each axis represents a task type.}
\label{fig:radar}
\end{figure}

Table~\ref{tab:difficulty} reports per-model accuracy by difficulty level; Figure~\ref{fig:difficulty} plots the trajectories, Figure~\ref{fig:radar} shows per-task accuracy profiles, and Figure~\ref{fig:correlation} gives the inter-task correlation structure.

The most striking case is LLaMA-3.3-70B, which scores \emph{higher} on hard items (90.6\%) than easy
ones (79.4\%), an 11.2-point inversion; five other models show the same higher-on-hard shape to a
lesser degree, and two show the opposite, a dip at medium difficulty with partial recovery on hard.
Neither shape is what ``harder'' was meant to produce. Hard items in this benchmark are typically
multi-instrument amendment-chain questions that state their cues explicitly in the text, whereas
medium items more often turn on a single subtle qualifying clause---a property closer to
\emph{ambiguity} than to \emph{reasoning depth}. A model with strong long-context tracking can find
an explicit multi-step chain easier to follow than a terse single clause that hinges on one word.

\begin{table}[t]
\centering\footnotesize
\begin{tabular}{lrrr}
\toprule
\textbf{Model} & \textbf{Easy} & \textbf{Med.} & \textbf{Hard} \\
 & \textit{(n=160)} & \textit{(n=182)} & \textit{(n=64)} \\
\midrule
Gemini 2.5 Flash & 92.5 & 89.0 & 84.4 \\
Qwen3-32B & 81.9 & 87.9 & 87.5 \\
LLaMA-3.3-70B & 79.4 & 85.2 & 90.6 \\
Llama 4 Scout 17B & 82.5 & 81.9 & 89.1 \\
Kimi K2 & 81.9 & 80.8 & 82.8 \\
LLaMA-3-8B & 76.2 & 79.7 & 78.1 \\
GPT-OSS 120B & 79.4 & 76.4 & 73.4 \\
GPT-OSS 20B & 75.0 & 79.7 & 73.4 \\
Gemini 2.5 Pro\textdagger{} & 83.1 & 72.5 & 68.8 \\
Mistral-7B & 74.4 & 76.9 & 76.6 \\
DeepSeek-R1-Distill-Llama-70B & 72.5 & 77.5 & 75.0 \\
Gemma 3 4B & 82.5 & 74.7 & 81.2 \\
\bottomrule
\end{tabular}
\caption{Accuracy (\%) by difficulty level (full 406-item evaluation, strict pipeline). \textdagger{}Gemini 2.5 Pro's decline is a scoring-regime artifact, not a measured verbosity effect; see Appendix~\ref{app:judge} and Section~\ref{sec:regime}.}
\label{tab:difficulty}
\end{table}

\section{Few-Shot Evaluation}
\label{app:fewshot}

To assess whether zero-shot evaluation represents a conservative lower bound on model capability, we evaluated four top-performing models---Gemini 2.5 Flash, LLaMA-3.3-70B, Llama 4 Scout 17B, and Qwen3-32B---under 3-shot prompting. For each task type, three fixed in-context examples were prepended to the prompt, drawn from the same distribution as the benchmark items and held constant across all models. All other evaluation conditions (system prompt, scoring pipeline, dataset) were identical to the zero-shot setup.

\begin{table*}[t]
\centering\small
\begin{tabular}{llrrrrr}
\toprule
\textbf{Model} & \textbf{Cond.} & \textbf{REG} & \textbf{NUM} & \textbf{CON} & \textbf{TMP} & \textbf{Overall} \\
\midrule
\multirow{3}{*}{Gemini 2.5 Flash} & 0-shot & 93.1\% & 84.8\% & 88.7\% & 88.5\% & 89.7\% \\
 & 3-shot & 92.0\% & 87.0\% & 91.9\% & 88.5\% & 90.1\% \\
 & $\Delta$ & $-$1.1 & +2.2 & +3.2 & +0.0 & +0.4 \\
\midrule
\multirow{3}{*}{Qwen3-32B} & 0-shot & 85.1\% & 77.2\% & 90.3\% & 92.3\% & 85.5\% \\
 & 3-shot & 86.8\% & 79.3\% & 90.3\% & 92.3\% & 86.7\% \\
 & $\Delta$ & +1.7 & +2.1 & +0.0 & +0.0 & +1.2 \\
\midrule
\multirow{3}{*}{LLaMA-3.3-70B} & 0-shot & 86.2\% & 75.0\% & 95.2\% & 79.5\% & 83.7\% \\
 & 3-shot & 87.4\% & 77.2\% & 98.4\% & 87.2\% & 86.7\% \\
 & $\Delta$ & +1.2 & +2.2 & +3.2 & +7.7 & +3.0 \\
\midrule
\multirow{3}{*}{Llama 4 Scout 17B} & 0-shot & 86.2\% & 66.3\% & 98.4\% & 84.6\% & 83.3\% \\
 & 3-shot & 90.2\% & 82.6\% & 85.5\% & 76.9\% & 85.2\% \\
 & $\Delta$ & +4.0 & +16.3 & $-$12.9 & $-$7.7 & +1.9 \\
\bottomrule
\end{tabular}
\caption{Zero-shot vs. 3-shot accuracy by task type.}
\label{tab:e1}
\end{table*}

3-shot prompting improves numerical reasoning by 2.1--16.3 percentage points across all four models, with Llama 4 Scout 17B showing the largest NUM gain (+16.3 pp, from 66.3\% to 82.6\%). LLaMA-3.3-70B gains +7.7 pp on temporal reasoning under 3-shot. These improvements are consistent with the direction of the zero-shot conservatism claimed in Section~\ref{sec:main-results-text}; with four models and one prompting configuration, this is suggestive rather than a controlled test of that claim.

CON performance is mixed: Llama 4 Scout 17B declines 12.9 pp (from near-ceiling 98.4\% to 85.5\%), while Gemini 2.5 Flash improves +3.2 pp. Performance degradation under 3-shot on tasks with near-ceiling zero-shot accuracy is consistent with in-context example interference, where the few-shot prefix biases the model away from clear Yes/No responses on straightforward contradiction items. Qwen3-32B is unaffected on CON and TMP (0.0 pp), suggesting its zero-shot behavior is already well-calibrated for these task types.

Overall, the broad performance tier ordering is preserved under 3-shot prompting (Gemini 2.5 Flash $\geq$ Qwen3-32B $\geq$ LLaMA-3.3-70B $\approx$ Llama 4 Scout 17B). We do not read this as establishing that the tier structure reflects genuine capability differences rather than prompting sensitivity: the same four models show a 12.9-point CON decline for Llama 4 Scout 17B under the identical 3-shot condition, which is itself a prompting-sensitivity effect. The evidence supports only the narrower claim that overall ordering is not obviously disrupted by this one alternative prompting configuration.

\end{document}